\documentclass{article}

\PassOptionsToPackage{numbers,compress}{natbib}
\usepackage[preprint]{neurips_2026}

\usepackage[utf8]{inputenc}
\usepackage[T1]{fontenc}
\usepackage{hyperref}
\usepackage{url}
\usepackage{booktabs}
\usepackage{amsfonts}
\usepackage{amssymb}
\usepackage{amsmath}
\usepackage{nicefrac}
\usepackage{microtype}
\usepackage{xcolor}
\usepackage{graphicx}
\usepackage{subcaption}          
\usepackage{multirow}
\usepackage{makecell}            
\usepackage[ruled,lined,noend]{algorithm2e}  
\usepackage{enumitem}            
\usepackage{wrapfig}
\usepackage{natbib}              
\usepackage{tcolorbox}           

\newcommand{\method}{\texttt{LeAct}}
\newcommand{\exit}{\textsc{ExIt}}
\renewcommand{\cot}{\textsc{CoT}}   
\newcommand{\nocot}{\textsc{NoCoT}}
\newcommand{\bc}{\textsc{BC}}
\newcommand{\bon}{\textsc{BoN}}
\newcommand{\kl}{\mathrm{KL}}

\title{\texttt{LeAct}: Learning to Reason from Expert Actions}

\author{
  Ziran Yang\thanks{Correspondence: \texttt{zirany@princeton.edu}}
  \quad Chengshuai Shi \quad Raj Ghugare \\
  \textbf{Benjamin Eysenbach \quad Karthik Narasimhan \quad Chi Jin} \\[0.35em]
  \textnormal{Princeton University}
}

\begin{document}
\maketitle

\begin{abstract}
Modern reasoning models depend on reasoning data, today sourced from human annotations or distilled from stronger LLMs.
However, a rich and largely untapped source of supervision lies in expert systems (e.g., game engines, classical planners, theorem provers), which routinely produce near-optimal actions across diverse domains.
But these experts are silent: they commit to an action without writing down the chain of thought (CoT) behind it.
Recovering that CoT as natural-language reasoning would distill expert knowledge into a student that generalizes beyond the demonstrated actions.
We treat it as a latent variable and study how to recover it from the action alone.
Our approach, \method{} (\textbf{Le}arning to reason from \textbf{Act}ions), optimizes this latent variable: 
the student samples candidate CoTs for each expert action, and we retain those that measurably improve its own probability of recovering the action.
Across imperfect-information games at multiple scales and a simulated robotics benchmark, \method{} reaches the solver's numerical floor on small enumerable games.
At larger scale, it is $5\times$ closer to the solver than the strongest expert-iteration baseline.
At Flop Hold'em ($\sim 10^9$ infosets), \method{} wins head-to-head by $+60$\,mbb/g, and on the robotics probe it is the only training recipe that improves on direct imitation.
We present a principled framework and the result: expert systems become a categorically new source of reasoning teachers for foundation models.

\end{abstract}

\section{Introduction}
\label{sec:introduction}

Building chain-of-thought (CoT) data is now a crucial bottleneck for reasoning language models~\citep{openai2024o1, guo2025deepseek}.
There is no shortage of work trying to scale beyond human annotation by automatically labeling CoT, either through distillation from stronger LLMs~\citep{ho2023large, magister2023teaching, guo2025deepseek} or by bootstrapping a model's own outputs~\citep{zelikman2022star, gulcehre2023reinforced, ruan2025reasoning}.
But these methods rarely close the loop directly on the CoT itself.
Filtering on synthetic CoT typically targets answer correctness or surface plausibility.
But whether the CoT actually contributes to the student's prediction remains uncertain.

We draw from a different source: silent action experts (game solvers~\citep{zinkevich2007regret,tammelin2014solving,brown2019deep}, classical planners, robotic planners~\citep{toussaint2015logic}, theorem provers~\citep{lin2025goedelv2}).
Each commits to a near-optimal action at every state, but neither the action nor the reasoning behind it is expressed in natural language.
The naive approach, asking a student LLM to write a CoT for each expert action, runs into rationalization~\citep{turpin2023language, lanham2023measuring}: the CoT can be plausible while bearing no causal link to the action it accompanies.
We close the loop instead.
For every state-action pair, the student samples multiple candidate CoTs, and we retain only those that raise the student's own probability of recovering the expert action.
The filter is the load-bearing step: replacing it with random ranking erases the advantage in our ablations.

\begin{figure}[!t]
  \centering
  \includegraphics[width=1.0\textwidth]{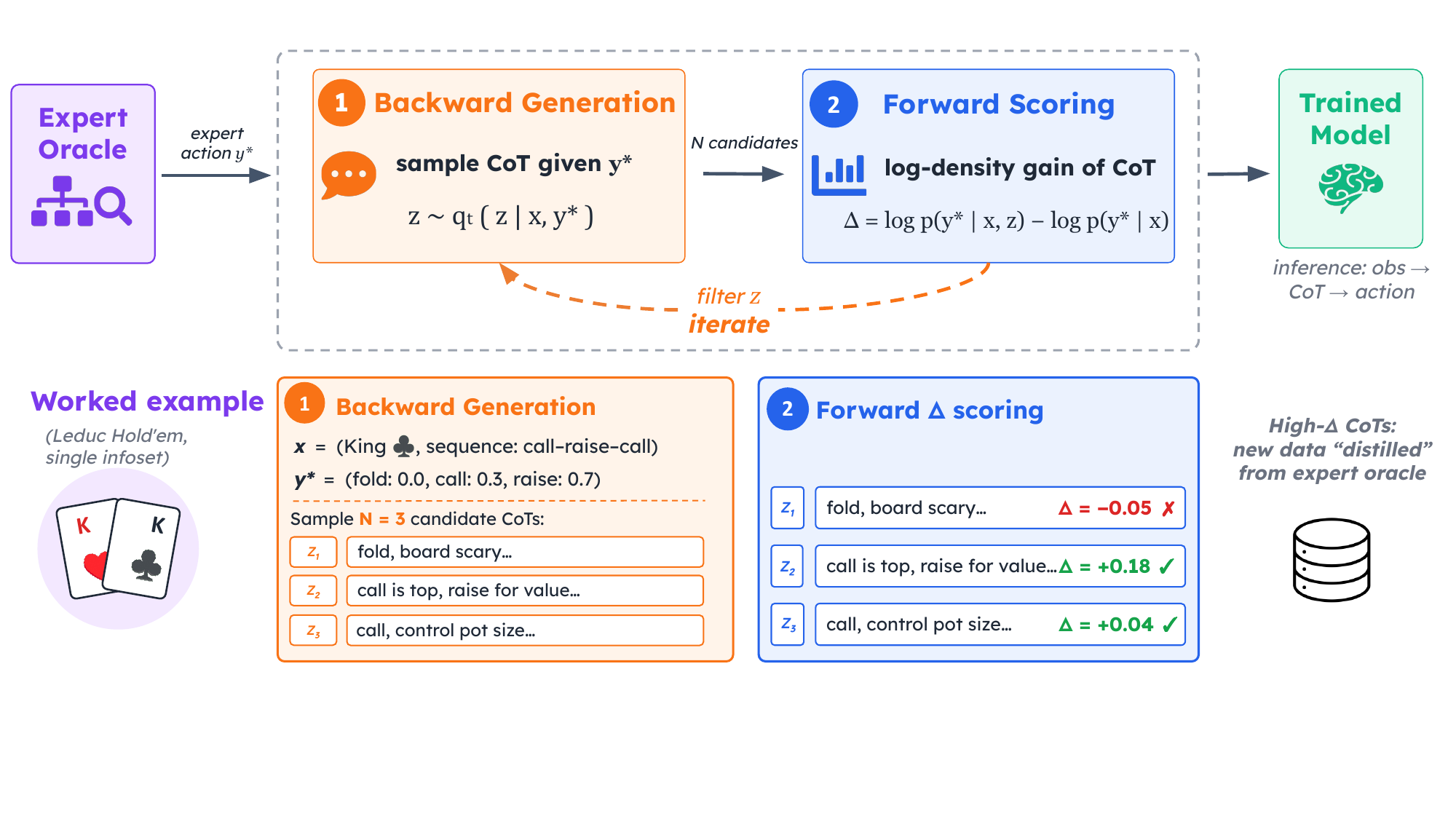}
  \vspace{-1.5cm}
  \caption{\textbf{\method{} turns silent action experts into reasoning teachers.}
    For each state $x$, the student samples candidate explanations conditioned on the oracle's action $\pi^*(\cdot \mid x)$.
    We keep those that raise the student's predicted probability of the action and use them as CoT to fine-tune the student.
    }
  \label{fig:overview}
  \vspace{-0.3cm}
\end{figure}

\method{} (\textbf{Le}arning to reason from \textbf{Act}ions) realizes this idea as an iterative training pipeline (Fig.~\ref{fig:overview}).
The conventional \emph{reason-then-act} paradigm~\citep{yao2023react} takes reasoning as input and produces actions at inference time;
\method{} runs that loop in reverse at training time, conditioning on the expert action, sampling reasoning that recovers it, and training on the reasoning that survived the filter.
The recipe accepts any action-only oracle: a CFR solver, a deep RL policy, or a frontier LLM committed to a single demonstration trajectory.
We test all three families: \texttt{CFR} solvers on enumerable poker settings like Leduc Hold'em~\citep{southey2005bayes}; 
a \texttt{DeepCFR}~\citep{brown2019deep} policy on Flop Hold'em ($\sim\!10^9$ infosets); 
and frontier-LLM successful trajectories on a simulated robotics benchmark~\citep{ghugare2025builderbench}.

\textbf{Contributions.}
We make three: a new non-LLM source of CoT supervision, a latent-variable derivation of the algorithm, and empirical results on various games and a robotics benchmark.

\begin{itemize}[leftmargin=*, topsep=2pt, itemsep=2pt]
  \item \textbf{A non-LLM data source for CoT supervision.}
  Rather than mining intermediate text~\citep{ruan2025reasoning,zelikman2022star}, \method{} converts a structured action policy $\pi^*(\cdot \mid x)$ from any non-LLM oracle into natural-language reasoning data, a categorically distinct supervision modality from prior CoT pipelines.

  \item \textbf{A principled algorithm} (§\ref{sec:method}).
  We treat the CoT as a latent variable connecting state $X$ to expert action $Y^* \sim \pi^*$, and derive a learning procedure from this view.
  The E-step reduces to ranking candidate CoTs by how much each helps the student's own decision-making, and the M-step prescribes $\pi^*$ as the SFT target.
  \method{} sits in the recent line of latent-EM self-improvement~\citep{ruan2025reasoning}, but anchors the latent to an external expert action rather than to text the model itself produced.

  \item \textbf{Empirical results} (§\ref{sec:results}--\ref{sec:analysis}). We test \method{} on imperfect-information games~\citep{lanctot2019openspiel} at multiple scales and on a simulated robotics cube-stacking benchmark~\citep{ghugare2025builderbench}.
  A student trained with \method{} matches the expert solver on small games, beats both behavior-cloning and expert-iteration baselines at the large games we test, and is the only recipe that improves on direct imitation in our robotics setting.
  The advantage holds on unseen states, and we trace it to the explanation-selection step.
\end{itemize}

\section{Related Work}
\label{sec:related_work}

\textbf{The closed-loop feedback gap in CoT supervision.}
There is no shortage of CoT-supervision work: humans label rationales by hand~\citep{wei2022chain, kojima2022large}, stronger LLMs distill them into smaller students~\citep{ho2023large, magister2023teaching, guo2025deepseek}, and self-improvement loops mine latent thoughts from text the model itself produces~\citep{zelikman2022star, hosseini2024vstar, zelikman2024quiet, gulcehre2023reinforced, singh2024beyond, ruan2025reasoning}.
What none of these provide is a closed-loop feedback signal that decides \emph{which} CoTs are worth training on by checking them against an external authority.
Existing pipelines either skip the filter entirely (annotation, distillation), filter by binary correctness against a single verifiable answer~\citep{zelikman2022star, gulcehre2023reinforced, singh2024beyond}, or filter by the model's own likelihood~\citep{ruan2025reasoning, burda2016iwae}.
In every case the supervision signal is either absent or LLM-internal.
\method{} fills this gap: each candidate CoT survives only if it measurably increases the student's own probability of recovering an external oracle's action, closing the loop with supervision that originates outside the LLM family.

\textbf{External verifiable feedback for training reasoning.}
A second line of work uses an external oracle or verifier as the training signal, instead of (or alongside) text the LLM itself produced.
Pairing a generator with an external checker is a long-running training recipe across game outcomes, formal verifiers, and theorem provers~\citep{silver2018general, ma2026evolving, shi2026odysseus, trinh2024alphageometry, hubert2025alphaproof, polu2020generative, lin2025goedel, lin2025goedelv2, zhao2026algoveri, li2026goedelcode, meta2022human}.
In the LLM-reasoning literature this signal takes three forms.
At the outcome level, reinforcement learning with verifiable rewards~\citep{shao2024deepseekmath, guo2025deepseek, lambert2024tulu3, openai2024o1, havrilla2024teaching, trung2024reft, kumar2024score, wang2024oreo} runs RL against a binary reward when the final answer matches ground truth.
Expert Iteration~\citep{anthony2017thinking} and rejection sampling fine-tuning~\citep{cobbe2021training, yuan2023scaling, touvron2023llama} use the same outcome signal in an SFT loop: sample candidates, keep those the oracle endorses, fine-tune on the survivors.
At the step level, process reward models~\citep{uesato2022solving, lightman2024lets, wang2023mathshepherd, zhang2024restmcts, setlur2024progress} train a separate verifier and apply it to each reasoning step.
\method{} uses a graded version of the Expert Iteration signal: each candidate CoT is scored by how much it raises the student's probability of the oracle's full action distribution, a continuous signal that works where the optimal action is mixed and no single token is the correct answer.
Conditioning the backward CoT on the expert action is a hindsight-relabelling step~\citep{andrychowicz2017hindsight, liu2024chain}, re-pairing each state with an outcome the model knows is correct.
A complementary line treats reasoning itself as inference-time search~\citep{hao2023reasoning, snell2024scaling}; \method{} acts at training time, selecting which CoT samples become supervision rather than how to deploy them.

\section{Method}
\label{sec:method}

The goal is to convert action-only oracles into thinking traces for training LLMs.
Two obstacles: the reasoning trace $z$ is never observed, inherently a latent variable; and the observed actions are in structured game action space rather than text, modality misalignment.
We design a closed-loop that adapts \citet{ruan2025reasoning}'s proposal-and-score framework to our scope of action supervision instead of text backfilling.

\subsection{Problem Formulation}
\label{sec:method_problem}

\textbf{Setup.} We consider a decision-making setting where states $x \in \mathcal{X}$ are drawn from a task of interest, in which an action oracle (a \texttt{CFR} solver, a classical planner, a verifiable controller, or a trained policy) supplies a per-state expert action distribution $\pi^*(\,\cdot \mid x)$.
The student LLM follows a reason-then-act paradigm, generating an intermediate chain-of-thought $z$ before committing to an action $y$;
we model this explicitly by treating $z$ as a latent reasoning variable and factoring the joint distribution as
$p_\theta(z,\, y \mid x) \;=\; p_\theta(z \mid x)\, p_\theta(y \mid x,\, z).$
Training data is $\{(x_i, \pi^*(\,\cdot \mid x_i))\}_i$: we observe states and expert policies but \emph{no} reasoning traces.

\textbf{Objective.} Maximise the marginal log-likelihood of expert actions under the student LLM,
\begin{equation}
  \mathcal{J}(\theta) \;=\; \mathbb{E}_{x}\,\mathbb{E}_{y \sim \pi^*(\cdot\mid x)}\!\bigl[\log p_\theta(y \mid x)\bigr],
  \label{eq:obj}
\end{equation}
intractable because $p_\theta(y \mid x) = \sum_z p_\theta(z \mid x)\, p_\theta(y \mid x, z)$ marginalises a latent natural-language trace.
The outer expectation $\mathbb{E}_{y \sim \pi^*}$ is realised at the M-step (§\ref{sec:method_training}): closed-form against $\pi^*$ when the oracle exposes a full distribution, and a Dirac mass when it commits to a single action.

\subsection{A View from Variational-EM}
\label{sec:method_em}

\textbf{Backward proposal and IWAE bound.} Introduce a backward proposal $q_\mathrm{bwd}(z \mid x, y)$, instantiated by prompting the LLM to generate a CoT given $(x, y)$, and draw $N$ candidates $z_1, \ldots, z_N \sim q_\mathrm{bwd}(\cdot\mid x, y)$ per state.
The $N$-sample importance-weighted bound~\citep{burda2016iwae}, which we denote $\mathcal{L}_N(\theta)$, is
\begin{equation}
  \mathcal{J}(\theta) \;\geq\; \mathcal{L}_N(\theta) \;:=\; \mathbb{E}_{x}\,\mathbb{E}_{y \sim \pi^*}\,\mathbb{E}_{z_{1..N} \sim q_\mathrm{bwd}(\cdot\mid x, y)}\!\left[\log\frac{1}{N}\sum_{i=1}^N \frac{p_\theta(z_i, y \mid x)}{q_\mathrm{bwd}(z_i \mid x, y)}\right],
  \label{eq:elbo}
\end{equation}
strictly tighter than the standard ELBO and exact as $N\!\to\!\infty$, holding for any $q_\mathrm{bwd}$ including a fixed LLM (which sidesteps the variational gradient through $q_\mathrm{bwd}$ that the $N{=}1$ ELBO would require).
\citet[Eq.~8]{burda2016iwae} show that the gradient of $\mathcal{L}_N(\theta)$ is a per-sample weighted SFT update,
\begin{equation}
  \nabla_\theta\,\mathcal{L}_N
  \;=\;
  \mathbb{E}_{x}\,\mathbb{E}_{y \sim \pi^*}\,\mathbb{E}_{z_{1..N} \sim q_\mathrm{bwd}}\!\left[\sum_{i=1}^N \widetilde w_i\, \nabla_\theta \log p_\theta(z_i, y \mid x)\right],
  \label{eq:iwae_grad}
\end{equation}
with normalised soft weights $\widetilde w_i = w_i / \sum_j w_j$ and $w_i = p_\theta(z_i, y \mid x) / q_\mathrm{bwd}(z_i \mid x, y)$, and $q_\mathrm{bwd}$ held fixed at the previous-round parameters within each round, evolving across rounds through the shared M-step updates.

\textbf{Factorising the IWAE weight $w_i$ into a task term and a proposal term.}
Bayes' rule factorises each per-sample weight $w_i$ from Eq.~\eqref{eq:iwae_grad} as
\begin{equation}
  w_i
  \;=\; \underbrace{\frac{p_\theta(y \mid x, z_i)}{p_\theta(y \mid x)}}_{=\;\exp\Delta(z_i;\, x, y)}
  \;\cdot\; \underbrace{\frac{p_\theta(z_i \mid x)}{q_\mathrm{bwd}(z_i \mid x, y)}}_{\rho(z_i;\, x, y)}
  \;\cdot\; p_\theta(y \mid x),
  \label{eq:isr}
\end{equation}
with $\Delta(z;\, x, y) := \log p_\theta(y \mid x, z) - \log p_\theta(y \mid x)$.
The marginal $p_\theta(y \mid x)$ is constant in $i$ and cancels in $\widetilde w_i = w_i / \sum_j w_j$.
The two remaining factors live on opposite sides of the action--CoT modality boundary.
$\Delta$ is a ratio over the structured expert action $y$ (a probability vector or discrete label whose task-level meaning is independent of its language rendering), and asks whether $z_i$ raises the student's probability of $y$ (implementation in §\ref{sec:method_scoring}).
$\rho$ is a ratio over the free-form CoT $z_i$, which has no such structured target; it measures whether the proposal exploited $y$ in \emph{generating} $z_i$, not whether $z_i$ usefully \emph{decides} $y$.
We therefore approximate $\widetilde w_i \propto \exp\Delta(z_i;\, x, y)$, retaining the task-grounded factor; dropping $\rho$ is unavailable to latent-text variants without an action anchor~\citep{ruan2025reasoning}.
§\ref{sec:analysis_memo_vs_reason} verifies the resulting $\Delta$-only filter is load-bearing; App.~\ref{app:em_derivation} gives the bound-level statement and the conditions under which the $\rho$-drop preserves ranking.

\textbf{LeAct objective.} Substituting $\widetilde w_i \propto e^{\Delta(z_i;\, x, y)}$ into Eq.~\eqref{eq:iwae_grad} yields the (approximate) M-step that defines the \method{} training objective:
\begin{equation}
  \boxed{\;\max_\theta\;\mathbb{E}_{x}\,\mathbb{E}_{y \sim \pi^*(\cdot\mid x)}\,\mathbb{E}_{z_{1..N} \sim q_\mathrm{bwd}}\!\Bigl[\, \sum_{i=1}^N \widetilde w_i\,\underbrace{\log p_\theta(z_i,\, y \mid x)}_{\text{SFT loss}}\,\Bigr],\quad \widetilde w_i \;=\; \frac{e^{\Delta(z_i;\, x, y)}}{\sum_{j=1}^N e^{\Delta(z_j;\, x, y)}}.\;}
  \label{eq:leact}
\end{equation}
In §\ref{sec:method_design} we approximate this softmax with a hard top-$K$ filter to recover a plain joint SFT loss.

\subsection{\method{} in Practice}
\label{sec:method_design}

We now specify the three stages that produce the SFT dataset $\mathcal{D}$ and realise the M-step: backward generation of CoT candidates, forward-delta evaluation with top-$K$ selection, and expert-policy forcing combined with joint backward supervision.

\textbf{Backward generation (E-step proposal).}
\label{sec:method_backward}
We instantiate the backward proposal $q_\mathrm{bwd}$ by prompting the current student.
The prompt gives the state $x$ and a qualitative summary $\tilde{\pi}^*$ of the expert distribution $\pi^*(\cdot \mid x)$ (for example, ``mostly fold with occasional calls''), and asks the student to explain why this strategy is optimal.
We draw $N$ candidates $z_1, \ldots, z_N$ per state.
The qualitative redaction prevents the backward generator from copying $\pi^*$ verbatim from its prompt (example traces in App.~\ref{app:examples}).
Conditioning on the expert action anchors the explanation in a known-good answer rather than searching forward for one, applying the hindsight-relabelling intuition (§\ref{sec:related_work}) to CoT supervision: the silent action becomes a teaching signal once paired with the explanation it licenses.

\textbf{Forward delta scoring and top-$K$ selection.}
\label{sec:method_scoring}
\label{sec:method_selection}
For each candidate $z_n$ we evaluate $\Delta(z_n; x, y)$ from Eq.~\eqref{eq:isr}, with both terms read off the student's parsed textual policy line over actions (implementation and fidelity audit in App.~\ref{app:textual_policy_proxy}).
The conditional $\log p_\theta(y \mid x, z_n)$ uses the line decoded under the candidate trace; the baseline $\log p_\theta(y \mid x) = \log \mathbb{E}_{z \sim p_\theta(z \mid x)}\bigl[p_\theta(y \mid x, z)\bigr]$ is estimated by Monte Carlo from $M$ self-generated forward traces $z^{(i)} \sim p_\theta(z \mid x)$ (no gold-action conditioning).
We then form the per-state average $\bar\Delta(z; x) = \mathbb{E}_{y \sim \pi^*}[\Delta(z; x, y)]$.
For oracles that commit to a single action rather than a distribution, the log-likelihood terms in $\Delta(z; x, y)$ are replaced with a heuristic action-match reward against $y^*$; the rest of the pipeline is unchanged.

Carrying the softmax weights $\widetilde w_i \propto e^{\Delta_i}$ into the M-step costs $O(N \cdot |\mathcal{X}|)$.
We sparsify in two complementary steps:
\begin{equation}
  \mathcal{Z}_x \;=\; \mathrm{top}\text{-}K\bigl\{z_j \sim q_\mathrm{bwd} : \bar\Delta(z_j;\,x) > 0\bigr\},
  \qquad
  \mathcal{D} \;=\; \{(x, z) : x \in \mathcal{X},\; z \in \mathcal{Z}_x\}.
  \label{eq:topk_selection}
\end{equation}
The positive-$\bar\Delta$ filter is a semantic admission criterion (only baseline-beating traces enter SFT), not a soft-weight approximation; it is what makes the round-over-round bound non-decreasing.
The top-$K$ indicator is the softmax truncation, cutting cost to $O(K \cdot |\mathcal{X}|)$.
The gate tightens across rounds: as the student improves, traces that merely match its current forward output drop out.

\textbf{Expert-policy forcing and joint backward supervision (M-step).}
\label{sec:method_training}
Applying the top-$K$ filter from Eq.~\eqref{eq:topk_selection} to the LeAct objective Eq.~\eqref{eq:leact} reduces the M-step to a joint SFT loss $\mathcal{L} = \mathcal{L}_\mathrm{fwd} + \mathcal{L}_\mathrm{bwd}$ on the selected traces $\mathcal{D}$.
$\mathcal{L}_\mathrm{fwd}$ trains the forward student on inputs $x$ to emit $z$ followed by the oracle's full action distribution $\pi^*(\cdot\mid x)$.
The analogy is teacher forcing in sequence modelling: there the ground-truth token replaces the model's own sample as the next-step target; here the oracle's distribution replaces any student-decoded policy as the action target.
We call this \emph{expert-policy forcing} (EPF).
$\mathcal{L}_\mathrm{bwd}$ retrains the backward proposal on inputs $(x, \tilde{\pi}^*)$ to emit $z$, the reweighted-wake-sleep~\citep{bornschein2014rwsleep} update on $q_\mathrm{bwd}$.
EPF is closed-form against a distributional oracle, and reduces to standard SFT on the single action when the oracle commits to one.
Across rounds, $\mathcal{L}_\mathrm{bwd}$ tracks $q_\mathrm{bwd}$ to the posterior under the current student, monotonically tightening the IWAE bound.
Full formulae are in App.~\ref{app:epf_derivation}; iteration dynamics are in App.~\ref{app:iteration_dynamics}.

\begin{algorithm}[t]
\small
  \caption{The \method{} algorithm}
  \label{alg:leact}
  \centering
  \begin{minipage}{0.9\linewidth}
    \DontPrintSemicolon
    \SetKwInOut{Input}{Input}
    \SetKwInOut{Output}{Output}
    \Input{Expert oracle $\pi^*$, initial model $\theta_0$, states $\mathcal{X}$, backward prompt template $P_\mathrm{bwd}$, candidates $N$, selection count $K$, rounds $R$}
    \Output{Trained model $\theta_R$}
    \For{round $r = 1, 2, \ldots, R$}{
      \tcp*[l]{E-step: backward generation}
      \For{each state $x \in \mathcal{X}$}{
        Sample $z_1, \ldots, z_N \sim p_{\theta_{r-1}}\!\bigl(\cdot \mid P_\mathrm{bwd}(x, \tilde{\pi}^*(x))\bigr)$\;
      }
      \tcp*[l]{E-step: forward delta scoring (impl.\ App.~\ref{app:textual_policy_proxy})}
      \For{each candidate $(x, z_j)$}{
        $\bar\Delta_j \leftarrow \bar\Delta(z_j;\, x)$ \tcp*[l]{average $\Delta$ from Eq.~\eqref{eq:isr}, $y \sim \pi^*(\cdot\mid x)$}
      }
      \tcp*[l]{Hard top-$K$ on positive deltas}
      \For{each state $x$}{
        $\mathcal{Z}_x \leftarrow \mathrm{top\text{-}}K\{z_j : \bar\Delta_j > 0\}$\;
      }
      \tcp*[l]{M-step: forward + backward joint SFT}
      $\mathcal{D}_r^\mathrm{fwd} \leftarrow \{(x,\;(z, \pi^*(\cdot\mid x))) : x \in \mathcal{X},\, z \in \mathcal{Z}_x\}$\;
      $\mathcal{D}_r^\mathrm{bwd} \leftarrow \{((x, \tilde{\pi}^*),\; z) : x \in \mathcal{X},\, z \in \mathcal{Z}_x\}$\;
      $\theta_r \leftarrow \mathrm{SFT}\bigl(\theta_{r-1},\; \mathcal{D}_r^\mathrm{fwd} \cup \mathcal{D}_r^\mathrm{bwd}\bigr)$\;
    }
    
    \Return{$\theta_R$}\;
  \vspace{-0.1cm}
  \end{minipage}
\end{algorithm}

\section{Experimental Setup}
\label{sec:experiments}

We evaluate \method{} on six imperfect-information games (Leduc Hold'em at three scales, Liar's Dice, 3-Player Leduc, Flop Hold'em) and the BuilderBench~\citep{ghugare2025builderbench} robotics cube-stacking benchmark, summarised in Table~\ref{tab:domains}; per-domain rules and oracle construction are in App.~\ref{app:domains}, \ref{app:oracle_construction}.

\textbf{Recipe and baselines.}
We use Qwen3-8B~\citep{qwen3} as the student LLM throughout.
Fixing the backbone is by design: it isolates supervision quality from model scale, Qwen3-8B is a representative open-weight student, and 8B fits within the compute budget required to fully train all seven settings.
The base model lacks the game-specific forward/backward competence to seed reasoning traces directly, so the two iterative methods (\cot{}\,\exit{} and \method{}) both initialise from a frontier-LLM-generated coldstart corpus and run 3 training rounds.
The coldstart is shared per-setting between the two, so any gap between them is attributable to the iteration phase.
Full setup details are in App.~\ref{app:details}.
\begin{itemize}[leftmargin=*, topsep=2pt, itemsep=2pt]
  \item \textbf{\nocot{}\,\bc{}} skips the CoT-candidate step: a single SFT pass on $(x, \pi^*(\cdot\mid x))$ pairs without a reasoning channel.
  Because it produces no CoT, it is not directly comparable on a data-centric basis with the iterative methods; we report it as a memorisation anchor.
  \item \textbf{\cot{}\,\exit{}} is standard expert iteration~\citep{anthony2017thinking,silver2018general} with a thinking model: each round samples $N{=}8$ forward rollouts $(z, y)$ per state, ranks them by KL of the model's parsed action distribution to the oracle, retains the top-$K{=}2$ as next-round SFT data, and uses the parsed action distribution itself (not $\pi^*$) as the action target (the standard expert-iteration practice; we ablate this in §\ref{sec:ablation_expert_policy}).
  \item \textbf{\method{}} samples $N{=}8$ backward CoT candidates per state conditioned on the expert action, ranks them by forward delta against the student's own self-baseline ($M{=}4$ parallel forward decodings, no gold-action conditioning), retains the top-$K{=}2$ as next-round SFT data, and uses $\pi^*$ as the action target (expert-policy forcing).
  This is not a design choice: the M-step prescribes it (§\ref{sec:method_em}), and empirically it is also the better target (§\ref{sec:ablation_expert_policy}).
\end{itemize}
Two reasons we do not implement RL/policy-gradient baselines (e.g., \texttt{GRPO}~\citep{guo2025deepseek}): this work targets SFT / mid-training, and \cot{}\,\exit{} already approximates a coarse-grained REINFORCE through its rollout-and-bootstrap loop.
Each round retains $K{=}2$ CoTs per state.
\cot{}\,\exit{} trains on the forward direction only ($K|\mathcal{X}|$ examples per round), while \method{}'s M-step jointly trains forward and backward directions ($2K|\mathcal{X}|$ per round).
For each (method, setting) we report at the round (out of 3) with the best evaluation metric.
Per-round dynamics, the stopping rule, and R1-only separation evidence are in App.~\ref{app:iteration_dynamics}.

\textbf{Two output formats.}
The seven settings split into two regimes by what the model emits at decoding time.
The six imperfect-information games (Leduc, Liar's Dice, 3-Player Leduc, FHP) ask the model to output a mixed-strategy \emph{distribution} per state.
Each forward decode emits a parsed action distribution whose KL to the oracle $\pi^*$ is the unit of supervision.
BuilderBench instead asks for a single JSON action per turn over a multi-turn cube-stacking rollout: success is multi-step progress.
Absolute scores are not comparable across regimes; only within-regime method ordering is.
Because BuilderBench's per-state KL is unavailable when the oracle commits to a single action, we replace forward-delta scoring (and \cot{}\,\exit{}'s ranking) with a heuristic action-match reward against the oracle's action $y^*$ for both methods (App.~\ref{app:builderbench_reward}); the rest of the recipe is unchanged.

\textbf{Metrics.}
For two-player zero-sum games (Leduc, Liar's Dice) we report \emph{exploitability} (best-response gap to Nash; $0$\,=\,Nash).
For 3-Player Leduc we report \texttt{NashConv}~\citep{lanctot2017unified}, an $n$-player generalisation that reduces to $2\times$ exploitability when $n{=}2$.
At FHP we report mean KL to the \texttt{DeepCFR} teacher and chip-outcome win rate in \emph{mbb/g} (milli-big-blinds per game, the standard heads-up play-strength unit; App.~\ref{app:fhp_winrate}).
On BuilderBench we report \emph{cube-placement progress} (per-task fraction of cubes placed at target positions, averaged over $64$ inference rollouts per task and then across tasks).
All reasoning-game metrics are reported under \textbf{single} (one CoT $\to$ one action) and \textbf{\bon{}} (per infoset, the candidate with smallest KL to $\pi^*$ across $N$ samples; oracle-aware), the standard inference-time scaling axis~\citep{wang2023selfconsistency}.
All are lower-is-better, except BuilderBench's cube-placement progress.

\textbf{Evaluation regime.}
Imitation is evaluated on the full game tree at enumerable scales (Leduc, Liar's Dice, 3-Player Leduc) and on BuilderBench's in-domain $26$-task partition (where the trajectory pool produced at least one fully-successful episode).
Generalisation is evaluated on a held-out infoset split at FHP, a feature-disjoint rank-split at Leduc 10r4s (test ranks never appear in training), and BuilderBench's OOD $20$-task partition (where the best-of-$3$ frontier-model trajectory pool produced no successful trajectory; App.~\ref{app:builderbench_partition}).

\begin{table}[t]
  \centering
  \small
  \caption{\textbf{Evaluation spans four orders of magnitude in scale across game families and a robotics domain.}
  Per-setting oracle, metric, and regime.}
  \label{tab:domains}
  \resizebox{\linewidth}{!}{%
  \begin{tabular}{lcccc}
    \toprule
    Domain & Scale & Oracle & Metric & Regime (split) \\
    \midrule
    Leduc 6r2s         & 4{,}032        & \texttt{CFR}             & Exploitability            & Imitation (full tree) \\
    Leduc 10r4s        & 47{,}040       & \texttt{CFR}             & Exploitability            & Both (full / rank-split) \\
    Leduc 13r4s        & 79{,}872       & \texttt{CFR}             & Exploitability            & Imitation (full tree) \\
    Liar's Dice        & 24{,}576       & \texttt{CFR}             & Exploitability            & Imitation (full tree) \\
    3-Player Leduc     & 13{,}878       & \texttt{CFR}             & \texttt{NashConv}         & Imitation (full tree) \\
    Flop Hold'em (FHP) & $\sim\!10^{9}$ & \texttt{DeepCFR}         & KL + mbb/g                & Generalisation (20K hold-out) \\
    BuilderBench       & 46 tasks       & Frontier-LLM traj.       & task progress & Both (in-domain / OOD) \\
    \bottomrule
  \end{tabular}}\\[2pt]
  {\scriptsize\itshape ``Both'' = the game is evaluated under both imitation (full tree) and generalisation (held-out split) regimes.}
  \vspace{-0.3cm}
\end{table}

\begin{figure}[t]
  \centering
  \includegraphics[width=.95\linewidth]{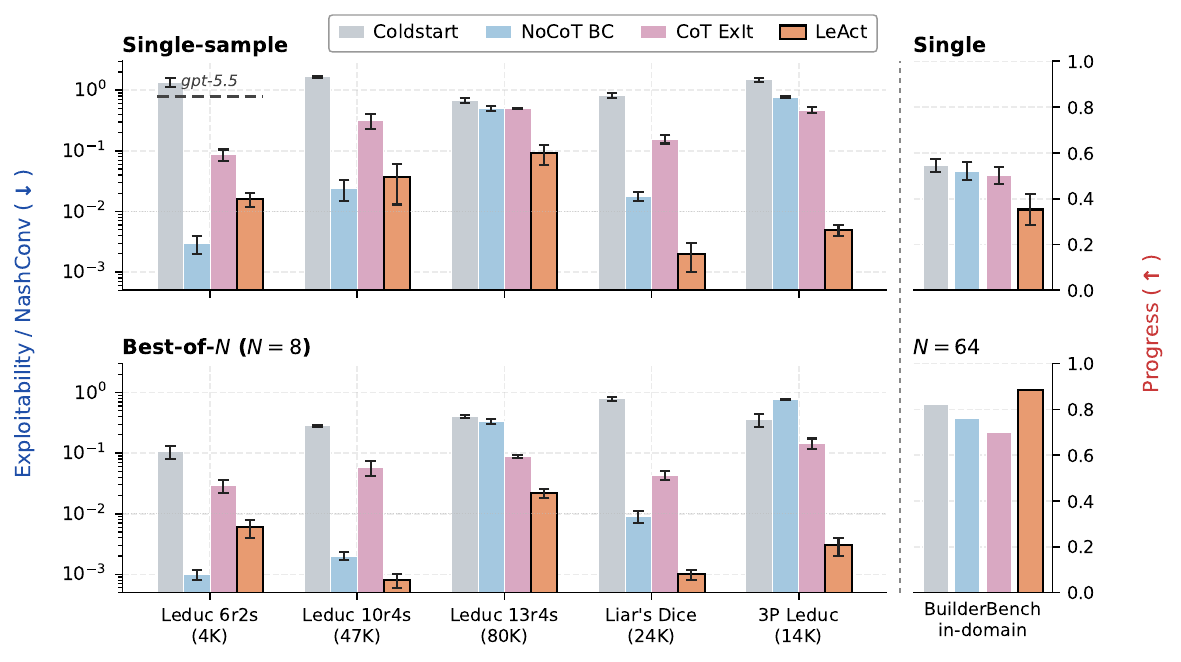}
  \caption{\textbf{Imitation quality: \method{} matches \nocot{}\,\bc{} where memorisation fits, and dominates where it doesn't.}
  Five game settings (4K--80K infosets, log $y$-axis, lower better,  \bon{} uses $N{=}8$) and BuilderBench in-domain 26 tasks (linear $y$-axis, higher better; $N{=}64$).
  Numbers in Table~\ref{tab:imitation}.
  }
  \label{fig:imitation}
  \vspace{-0.50cm}
\end{figure}

\section{Results}
\label{sec:results}

\method{} treats expert actions as imitation supervision while producing a latent CoT alongside each action; the empirical question is whether this latent helps relative to imitation without latent (\nocot{}\,\bc{}) or with self-generated latent that bypasses our backward-and-score loop (\cot{}\,\exit{}).
We organise the comparison along two axes.
\textbf{Imitation quality} (§\ref{sec:results_main}): under matched coldstart and data budget, how closely does the trained forward policy track the expert?
We measure across enumerable game scales (Leduc $4$K--$80$K), player counts (3-player Leduc), game families (Liar's Dice), and domains (BuilderBench cube-stacking).
\textbf{Generalisation} (§\ref{sec:results_generalization}): does the advantage persist on states the model never saw during training, where memorisation alone would not transfer?
We test on the Flop Hold'em hold-out infoset, the Leduc rank-split, and BuilderBench's OOD task partition.

\subsection{Imitation}
\label{sec:results_main}

\method{} matches or beats both baselines on \bon{} inference everywhere outside the 6r2s memorisation regime (Fig.~\ref{fig:imitation}).
The \bon{} advantage comes from generation diversity rather than higher single-sample quality: at Leduc 10r4s, \method{}'s single-to-\bon{} ratio is several times that of \nocot{}'s, and on BuilderBench in-domain \method{} starts at the lowest single-sample progress but leads at \bon{} across all four recipes.
The CoT spreads the per-state action distribution, so \bon{} catches better samples than any single draw.

\textbf{Scaling on poker.}
At 6r2s ($4$K infosets) the Nash table fits in model capacity and \nocot{}'s tabular memorisation suffices.
At 13r4s ($80$K infosets) memorisation breaks: \nocot{}\,\bc{} also stalls (App.~\ref{app:imitation_detail}), so the matched \cot{}\,\exit{} ablation (same coldstart, same supervision, same data (game states) for training) stalling $5.4\times$ worse on single-sample and $4\times$ worse on \bon{} than \method{} isolates backward-delta selection as the load-bearing component, not the presence of a CoT itself.
The trend is consistent: as scale grows, the value of selecting which CoTs to learn from grows.

\textbf{Cross-domain results.}
Beyond two-player Leduc the same imitation gap appears in three further regimes (different game family, multi-player, robotics).
On Liar's Dice (different game family), \method{} lands within solver-evaluation noise of the \texttt{CFR} oracle.
On 3-player Leduc, \method{} lands within solver noise while \exit{} stalls well above it.
At $n\!\geq\!3$ players the 2P-zero-sum convergence guarantee for best-response iteration fails (Banach contraction~\citep{banach1922operations}), so \exit{}'s table-fit policies have no monotone-improvement guarantee across rounds; only \method{}'s reasoning-based policy stays self-consistent across roles.
On BuilderBench cube-stacking in-domain set (using a frontier-LLM successful trajectory pool as oracle), \method{} is the only training recipe whose \bon{} progress clears the joint coldstart anchor; the matched \cot{}\,\exit{} comparator even regresses below it (Table~\ref{tab:imitation}).

Absolute KL and exploitability values are not directly comparable across these games: action-space size and Nash-policy concentration both differ, and large game trees like FHP have many infosets where the optimal move is near-deterministic, pulling the average KL floor lower.
What is interpretable across rows is therefore the within-game ordering rather than the absolute scale.

\begin{table}[t]
\centering
\small
\caption{\textbf{\method{} generalises across three held-out regimes.}
(a) KL ($\downarrow$) on a random infoset hold-out (FHP) and a feature-disjoint rank-split (Leduc 10r4s, test ranks never appear in training); cube-placement progress ($\uparrow$) on the BuilderBench (BB) OOD partition.
Two frontier-anchor rows: gpt-5.5 pass@$1$ on FHP (a comparator, not data source); the best-of-$3$ frontier-LLM trajectory pool used as BuilderBench SFT data, evaluated as a strength baseline on the same OOD partition.
(b) FHP chip outcome in mbb/g over $50$K paired hands (head-to-head; $\pm$ standard error).
Details in App.~\ref{app:fhp_winrate},~\ref{app:builderbench_sft_headroom}.}
\label{tab:generalisation_combined}

\begin{minipage}[t]{0.57\linewidth}
\vspace{0pt}
\begin{subtable}[t]{\linewidth}
\centering
\caption{Held-out generalisation.}
\label{tab:generalisation}
\setlength{\tabcolsep}{4pt}
\resizebox{\linewidth}{!}{%
\begin{tabular}{lccccc}
  \toprule
   & \multicolumn{2}{c}{Leduc 10r4s ($\downarrow$)} & \multicolumn{2}{c}{FHP $20$K ($\downarrow$)} & BB OOD ($\uparrow$) \\
  \cmidrule(lr){2-3} \cmidrule(lr){4-5} \cmidrule(lr){6-6}
  Method & Single & \bon{} & Single & \bon{} & Progress \\
  \midrule
  gpt-5.5  & ---             & ---             & 0.7530          & ---              & ---              \\
  Frontier traj. & ---             & ---             & ---             & ---              & 0.4923           \\
  Coldstart            & 2.462           & 0.332           & 0.0321          & 0.0018           & 0.4815           \\
  \nocot{}\,\bc{}    & 2.110           & 0.574           & 0.0145          & 0.0049           & 0.3980           \\
  \cot{}\,\exit{}    & 2.071           & 0.459           & 0.0053          & 0.0011           & 0.4361           \\
  \method{}         & \textbf{0.716}  & \textbf{0.168}  & \textbf{0.0019} & \textbf{0.0005}  & \textbf{0.5777}  \\
  \bottomrule
\end{tabular}}
\end{subtable}
\end{minipage}\hfill
\begin{minipage}[t]{0.39\linewidth}
\vspace{0pt}
\begin{subtable}[t]{\linewidth}
\centering
\caption{FHP chip outcome.}
\label{tab:fhp_winrate}
\setlength{\tabcolsep}{3pt}
\resizebox{\linewidth}{!}{%
\begin{tabular}{lccc}
  \toprule
   & \nocot{} & \cot{} & \method{} \\
  \midrule
  \nocot{}   & ---                              & $+41$\,{\tiny$\pm31$}  & $-60$\,{\tiny$\pm13$}  \\
  \cot{}     & $-41$\,{\tiny$\pm31$}      & ---                          & $-62$\,{\tiny$\pm28$}  \\
  \method{}  & $+60$\,{\tiny$\pm13$}      & $+62$\,{\tiny$\pm28$}  & ---                          \\
  \bottomrule
\end{tabular}}
\end{subtable}
\end{minipage}
\vspace{-0.25cm}
\end{table}

\subsection{Generalisation}
\label{sec:results_generalization}
\label{sec:results_scaling}

\method{} carries its advantage to states the model never saw during training, across three regimes: a random infoset hold-out at million-scale (FHP), a feature-disjoint rank-split at medium scale (Leduc 10r4s), and a task-level OOD partition on BuilderBench.
Random hold-outs test that the trained policy is not memorising the training infosets; feature-disjoint splits stress transfer along an axis the model never saw; the BuilderBench OOD partition stresses generalisation beyond the supervised demonstration distribution.

\textbf{Random hold-out at scale: FHP.}
Flop Hold'em has on the order of $10^9$ playable infosets, several orders of magnitude beyond the $80$K training infosets, so even random sampling exposes the policy to a state distribution it has never been trained on, making FHP a natural generalisation testbed even under the random hold-out regime.
The absolute KL floor is small because most infosets have near-trivial optimal strategy (preflop fold lines on weak hands).
The within-regime ordering is nonetheless sharp: \method{} beats both baselines, and a frontier zero-shot anchor (gpt-5.5; Table~\ref{tab:generalisation}) lands orders of magnitude worse.
Even a strong general reasoner is therefore severely under-optimised at producing calibrated numerical mixed strategies for a fine-grained game.
Chip-outcome head-to-head (Table~\ref{tab:fhp_winrate}) corroborates the KL ordering.

\textbf{Feature-disjoint shift: Leduc 10r4s rank-split.}
The rank-split (train ranks $\{2,4,6,8,\text{T}\}$, test $\{3,5,7,9,\text{J}\}$) forces the model to transfer along a feature axis it never saw during training, a stronger distribution-shift test than random hold-out.
All three methods are retrained from scratch for this split: the joint coldstart corpus is regenerated on the train ranks alone, and \cot{}\,\exit{} and \method{} then iterate from this rank-restricted coldstart base.
The test ranks $\{3,5,7,9,\text{J}\}$ are never seen during coldstart or any iteration round.
\method{}'s ${\sim}3\times$ lead over \cot{}\,\exit{} on the held-out ranks (Table~\ref{tab:generalisation}) shows that the imitation gap is not memorisation residue, though absolute values remain an order of magnitude above the imitation regime (single $0.716$ vs $0.037$): the gain reduces catastrophic generalisation failure rather than achieving clean transfer.

\textbf{Task-level OOD on BuilderBench.}
The OOD partition is $20$ tasks that no frontier model in the best-of-$3$ trajectory pool fully solved, so the SFT data contains no golden trajectory for any of these.
We report \emph{cube-placement progress}, a partial-solve metric (fraction of cubes correctly placed), so non-zero scores correspond to placing some cubes correctly even when the task is not solved end-to-end.
\method{} leads on this partition (Table~\ref{tab:generalisation}), and the gain is broad-based across tasks rather than concentrated on a few easy ones, so the model makes measurable progress on tasks it had no demonstration for.

\section{Analysis}
\label{sec:analysis}

§\ref{sec:results_main} establishes that backward-delta selection drives \method{}'s gap over the matched \cot{}\,\exit{} comparator.
This section runs four probes that test whether the gap might instead come from one of four alternatives, and closes with a compute-overhead analysis.
(1) Could any ranking rule have done equally well?
(2) Does our parsed-string proxy faithfully measure the formal $\Delta$?
(3) Is the gain just memorising surface features (e.g. Nash numbers)?
(4) Or is it the supervision-target choice (not backward delta) doing the work?

\begin{figure}[!t]
\centering
\begin{minipage}[t]{0.58\linewidth}
  \centering
  \vspace{0pt}
  \includegraphics[width=\linewidth]{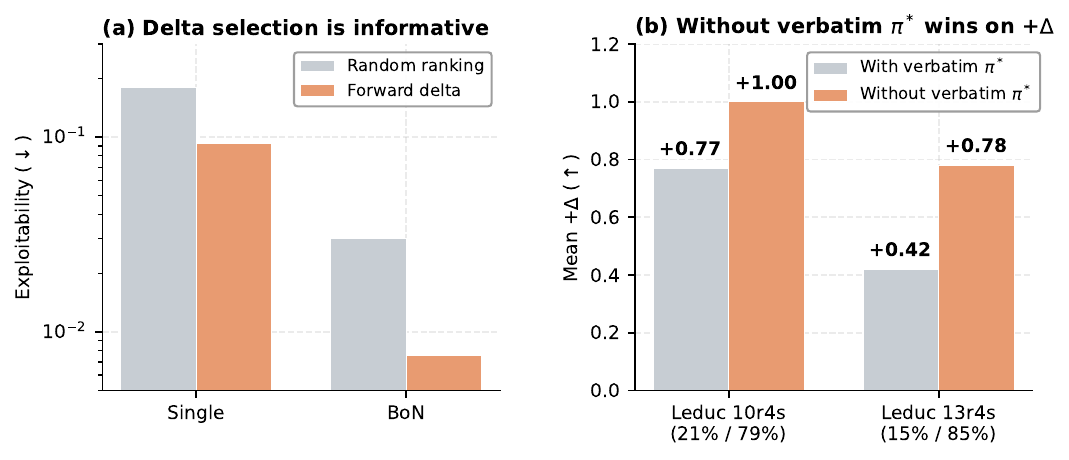}
  \captionof{figure}{\textbf{Two falsifiers of the shortcut hypothesis for backward-delta selection.}
  (a) Forward-delta vs.\ random ranking on Single and \bon{} (Leduc 10r4s).
  (b) Mean $+\Delta$ by verbatim-$\pi^*$ status.}
  \label{fig:e_step_ablations}
\end{minipage}\hfill
\begin{minipage}[t]{0.40\linewidth}
  \centering
  \vspace{0pt}
  \resizebox{\linewidth}{!}{%
  \begin{tabular}{llcc}
    \toprule
    Method & Regime & Single ($\downarrow$) & \bon{} ($\downarrow$) \\
    \midrule
    Coldstart   & ---    & 1.6701          & 0.2824           \\
    \cot{}\,\exit{} & EPF    & 1.0278          & 0.6191           \\
    \cot{}\,\exit{} & no-EPF & 0.4432          & 0.0599           \\
    \method{}   & EPF    & \textbf{0.1566} & \textbf{0.0019} \\
    \method{}   & no-EPF & 1.0565          & 0.1940           \\
    \bottomrule
  \end{tabular}}
  \captionof{table}{\textbf{Method and supervision target are entangled}
  We cross method (\cot{}\,\exit{} vs \method{}) with \emph{expert-policy forcing} (use the oracle's $\pi^*$ vs the model's own action) on Leduc \texttt{10r4s}.
  Matched pairings: \method{}+EPF and \cot{}\,\exit{}+no-EPF.
  }
  \label{tab:expert_policy_ablation}
\end{minipage}
\vspace{-0.45cm}
\end{figure}

\textbf{Delta selection is informative, not random.}
\label{sec:analysis_memo_vs_reason}
\label{sec:analysis_cot}
To test this, we replace forward-delta ranking with random ranking on Leduc 10r4s, sharpened to top-$1$ (so any gap comes from the ranking rule, not the data budget).
Forward-delta beats random on both single and \bon{} exploitability (Fig.~\ref{fig:e_step_ablations}a).
\method{}'s gain therefore comes from \emph{which} CoTs we keep, not just from keeping any $K=2$ subset.

\textbf{The textual proxy faithfully reads the formal $\Delta$.}
\method{} scores CoT candidates by reading the model's parsed policy distribution, not by computing the token-level log-likelihoods that the formal $\Delta$ is defined over.
This is a proxy.
We check whether it ranks the same way the formal $\Delta$ would: on Leduc 13r4s, the parsed-string $\bar\Delta$ and a token-level estimator $\Delta_{\mathrm{ll}}$ agree on the top-$1$ CoT in $89$--$98\%$ of states (App.~\ref{app:textual_policy_proxy}, Table~\ref{tab:sfa}).

\textbf{The selection signal is not memorised Nash numbers.}
If \method{}'s scorer were just rewarding CoTs that quote oracle probabilities verbatim (e.g., ``fold $0.376$''), then CoTs without such verbatim numbers would score worse.
We split the backward pool by verbatim-$\pi^*$ presence: at both Leduc 10r4s and 13r4s, the larger ``no verbatim'' stratum has a higher mean positive delta than the smaller ``with verbatim'' stratum (Fig.~\ref{fig:e_step_ablations}b).
The held-out generalisation results (§\ref{sec:results_generalization}) corroborate: a memorisation shortcut would not transfer to unseen states.

\textbf{The gap is backward delta, not the supervision target.}
\label{sec:ablation_expert_policy}
To separate target choice from method choice, we run a $2\times 2$ crossing method (\cot{}\,\exit{} vs \method{}) with target (\emph{expert-policy forcing} or EPF, vs the model's own action).
Each cell is a separate run from the shared Leduc 10r4s coldstart, iterated for 2 rounds at matched data budget.
The principled pairs are the diagonals: \method{}+EPF (the M-step prescribes $\pi^*$) and \cot{}\,\exit{}+no-EPF (standard ExIt does not relabel actions).
Only these diagonal cells beat coldstart (Table~\ref{tab:expert_policy_ablation}).
The off-diagonals collapse because in poker the CoT's textual policy line and the model's decoded action are essentially the same: supervising one with the other is self-referential and provides no new signal (App.~\ref{app:coherence_diagnostic}).
Within the diagonal, \method{}+EPF dominates \cot{}\,\exit{}+no-EPF on both Single ($0.16$ vs $0.44$) and \bon{} ($0.0019$ vs $0.06$), so backward-delta scoring carries the gap, not the EPF target.
The cross-domain BuilderBench replication breaks this coupling and recovers a clean ordering on both axes (App.~\ref{app:builderbench_sft_headroom}).

\textbf{Compute overhead.}
\method{} and \cot{}\,\exit{} both generate $N$ CoTs per state during training (backward for \method{}, forward for \cot{}\,\exit{}); \method{}'s only extra cost is $M$ parallel short action generation for forward scoring, negligible relative to the shared long CoT generation pass.
The two methods therefore incur essentially the same training compute.

\section{Conclusion}
\label{sec:conclusion}

The key idea of this paper is that many domains already have a silent domain-specific expert, and that the reasoning trace these experts cannot articulate can be recovered by the student itself and filtered against the expert's own actions.
This opens a new data source for reasoning supervision that does not require a human annotator or a stronger LLM, and the oracle itself need not be exact.
At Flop Hold'em we score against a \texttt{DeepCFR} teacher, and on BuilderBench against a pool of frontier-model trajectories, in both cases without modification to the recipe.

Other silent experts (protein structure predictors, formal proof checkers, robotic-control libraries) are natural candidates for the same recipe, and the most direct algorithmic extension is merging \method{} with online RL, where the oracle's action distribution can serve as a process-level baseline for credit assignment.
How \method{} interacts with frontier-scale students, beyond the 8B fixed here, is another direct extension.

\clearpage

\bibliographystyle{plainnat}
\bibliography{references}

\appendix

\clearpage

\section{Formulation: Derivations}
\label{app:formulation_derivations}

This appendix gives the derivations summarised in §\ref{sec:method_em}: (i) the IWAE bound, per-sample importance-weight factorisation, and rho-dropped soft-weight approximation that define the \method{} training objective (App.~\ref{app:em_derivation}); (ii) the M-step target-choice derivation for expert-policy forcing (App.~\ref{app:epf_derivation}); (iii) the correspondence between formal quantities and \method{} algorithm components (App.~\ref{app:em_correspondence}); and (iv) the textual policy proxy that bridges the formal $\Delta$ to its parsed-policy implementation (App.~\ref{app:textual_policy_proxy}).

\subsection{From the ELBO to the LeAct Objective}
\label{app:em_derivation}

This appendix expands §\ref{sec:method_em} line by line.
The derivation specialises the importance-weighted variational bound (IWAE) of \citet{burda2016iwae} to a setting in which the proposal $q_\mathrm{bwd}(z \mid x, y)$ is a fixed (within a round) LLM that conditions on $(x, y)$, and in which two of the four log-likelihood terms in the importance weight are intractable under a decoder-only model.
The same IWAE bound was recently applied to latent reasoning data synthesis on unlabelled text corpora by \citet{ruan2025reasoning}; the present derivation differs in bounding the conditional likelihood $\log p_\theta(y \mid x)$ for an expert-anchored target $y \sim \pi^*$ rather than the corpus marginal $\log p(X)$, which enables the additional reduction of the IWAE soft weight to a forward delta (Step~3).
The argument proceeds in four steps: (i) introduce the backward proposal and obtain an ELBO valid for any choice of $q_\mathrm{bwd}$, including a fixed LLM; (ii) factorise the importance ratio via Bayes' rule; (iii) keep only the two computable terms, yielding $\Delta$; (iv) re-weight the ELBO by $e^\Delta$ and replace the soft weighting by a hard top-$K$ filter for tractable SFT.

\textbf{Step 1: IWAE bound via a backward proposal.}
The marginal log-likelihood of an expert action $y \sim \pi^*(\,\cdot \mid x)$ under the student is $\log p_\theta(y \mid x) = \log \sum_z p_\theta(z \mid x)\, p_\theta(y \mid x, z)$, intractable over natural-language traces.
Introduce a backward proposal $q_\mathrm{bwd}(z \mid x, y)$ (instantiated by prompting the LLM to generate a CoT given $(x, y)$) and draw $N$ candidates $z_1, \ldots, z_N \sim q_\mathrm{bwd}(\cdot \mid x, y)$ i.i.d.
The $N$-sample importance-weighted bound of \citet{burda2016iwae} is
\begin{equation}
  \log p_\theta(y \mid x)
  \;\geq\;
  \mathbb{E}_{z_{1..N} \sim q_\mathrm{bwd}}\!\left[\log\frac{1}{N}\sum_{i=1}^N \frac{p_\theta(z_i \mid x)\, p_\theta(y \mid x, z_i)}{q_\mathrm{bwd}(z_i \mid x, y)}\right],
  \label{eq:app_elbo_per_xy}
\end{equation}
strictly tighter than the standard ELBO at $N{=}1$ and exact in the limit $N\!\to\!\infty$.
Taking the outer expectation over $x$ and $y \sim \pi^*$ gives the population IWAE bound,
\begin{equation}
  \mathcal{J}(\theta)
  \;\geq\;
  \mathbb{E}_{x}\,\mathbb{E}_{y \sim \pi^*}\,\mathbb{E}_{z_{1..N} \sim q_\mathrm{bwd}}\!\left[\log\frac{1}{N}\sum_{i=1}^N \frac{p_\theta(z_i, y \mid x)}{q_\mathrm{bwd}(z_i \mid x, y)}\right].
  \label{eq:app_elbo}
\end{equation}
\citet{burda2016iwae} show (their Eq.~8) that the gradient of Eq.~\eqref{eq:app_elbo} reduces to a per-sample SFT update with normalised soft weights,
\begin{equation}
  \nabla_\theta \mathcal{L}_N
  \;=\;
  \mathbb{E}_{z_{1..N}}\!\left[\sum_{i=1}^N \widetilde w_i\, \nabla_\theta \log p_\theta(z_i, y \mid x)\right],
  \quad
  \widetilde w_i = \frac{w_i}{\sum_{j=1}^N w_j},\quad w_i = \frac{p_\theta(z_i, y \mid x)}{q_\mathrm{bwd}(z_i \mid x, y)},
  \label{eq:app_iwae_grad}
\end{equation}
with $q_\mathrm{bwd}$ held fixed at $\theta_{r-1}$ within the round (so $\nabla_\theta \log q_\mathrm{bwd} = 0$ for the per-round bound); the inner term $\log p_\theta(z_i, y \mid x)$ is exactly the joint SFT loss on $(x, z_i, y)$.

The lower bound in Eq.~\eqref{eq:app_elbo} is valid for any choice of $q_\mathrm{bwd}(z \mid x, y)$, including a fixed LLM, and $N > 1$ tightens it strictly over the $N{=}1$ ELBO without optimising $q_\mathrm{bwd}$'s parameters.
This sidesteps the standard variational-inference gradient on $q_\mathrm{bwd}$ (expensive when $q_\mathrm{bwd}$ is an LLM) that would otherwise be required to tighten the ELBO further.
The per-sample weights $w_i = p_\theta(z_i, y \mid x) / q_\mathrm{bwd}(z_i \mid x, y)$ inside the log-mean are importance weights between the joint $p_\theta(z, y \mid x)$ and the proposal $q_\mathrm{bwd}(z \mid x, y)$; Bayes' rule factorises them in Step~2.
Within each \method{} round, $q_\mathrm{bwd}$ is held fixed at $\theta_{r-1}$; across rounds it evolves alongside the shared model parameters that the M-step trains, recovering an approximate-EM alternation (the correspondence is summarised in App.~\ref{app:em_correspondence}).
The practical pipeline additionally augments the M-step with an explicit backward SFT term (§\ref{sec:method_training}), which arises as the reverse-KL update on $q_\mathrm{bwd}$ in the reweighted-wake-sleep regime; Step~5 below derives this.

\textbf{Step 2: per-sample importance-weight factorisation.}
For each sample $z_i$ with $q_\mathrm{bwd}(z_i \mid x, y) > 0$, Bayes' rule factorises the unnormalised IWAE weight $w_i = p_\theta(z_i, y \mid x) / q_\mathrm{bwd}(z_i \mid x, y)$ as
\begin{equation}
  w_i
  \;=\;
  \underbrace{\frac{p_\theta(y \mid x, z_i)}{p_\theta(y \mid x)}}_{=\,\exp\Delta(z_i;\,x,\,y)}
  \;\cdot\;
  \underbrace{\frac{p_\theta(z_i \mid x)}{q_\mathrm{bwd}(z_i \mid x, y)}}_{\rho(z_i;\,x,\,y)}
  \;\cdot\;
  p_\theta(y \mid x),
  \label{eq:app_isr}
\end{equation}
with $\Delta(z;\, x, y) := \log p_\theta(y \mid x, z) - \log p_\theta(y \mid x)$.
The factor $p_\theta(y \mid x)$ does not depend on $i$ and so cancels in the normalised weight $\widetilde w_i = w_i / \sum_j w_j$; only $\exp\Delta_i$ and $\rho_i$ remain. The first factor $\exp\Delta_i$ is computable from the student's own logits: $\log p_\theta(y \mid x, z_i)$ is a teacher-forced log-likelihood of the action under the student conditioned on $z_i$, and $\log p_\theta(y \mid x) = \log \mathbb{E}_{z' \sim p_\theta(\cdot\mid x)}[p_\theta(y \mid x, z')]$ is estimable by Monte Carlo from a small number of forward samples (§\ref{sec:method_scoring}).

\textbf{Step 3: keeping the computable terms in the soft weight.}
Dropping the $i$-independent $p_\theta(y \mid x)$ that cancels in $\widetilde w_i$, the log per-sample weight admits the additive decomposition
\[
\log w_i
\;\propto\;
\underbrace{\log p_\theta(y \mid x, z_i) - \log p_\theta(y \mid x)}_{\Delta(z_i;\,x,\,y)}
\;+\;
\underbrace{\log p_\theta(z_i \mid x) - \log q_\mathrm{bwd}(z_i \mid x, y)}_{\log \rho(z_i;\,x,\,y)} .
\]
The first two terms (forming $\Delta$) are computable from the student's logits: $\log p_\theta(y \mid x, z_i)$ is a teacher-forced action log-likelihood, and $\log p_\theta(y \mid x) = \log \mathbb{E}_{z' \sim p_\theta(\cdot\mid x)}[p_\theta(y \mid x, z')]$ is estimable by Monte Carlo from a small number of forward samples (§\ref{sec:method_scoring}).
In practice the last two terms are difficult to compute and we omit them.
\method{} therefore approximates the unnormalised IWAE soft weight by
\begin{equation}
  w_\mathrm{LeAct}(z_i;\, x, y) \;:=\; \exp\Delta(z_i;\, x, y),
  \label{eq:app_w}
\end{equation}
i.e.\ replaces the soft weight $\widetilde w_i \propto e^{\Delta_i} \cdot \rho_i$ by the rho-dropped surrogate $\widetilde w^\mathrm{LeAct}_i \propto e^{\Delta_i}$.
This $\rho$-drop is the design choice the expert-anchored setting enables: without an external action target (e.g., latent-text variational EM~\citep{ruan2025reasoning}), the IWAE weight has no separable $\Delta$ and $\rho$ remains entangled with the bound.
§\ref{sec:analysis_memo_vs_reason} verifies empirically that the resulting $\Delta$-only filter is load-bearing.

\textbf{When dropping $\rho$ is safe.}
The action anchor in the backward prompt fixes the conditioned target $y$ across all candidates, so $\rho_i$ varies with $z_i$ only through how well each candidate trace explains the same fixed $y$.
For high-probability oracle actions, $\rho_i$ concentrates and acts approximately as a constant scaling factor across candidates, so the ranking induced by $e^{\Delta_i}$ matches the ranking induced by $e^{\Delta_i}\rho_i$ up to ties.
The drop biases ranking only when $\rho_i$ correlates with $\Delta_i$ across candidates, which would require the backward proposal to systematically place mass on high-$\Delta$ traces in a way that cancels with the forward likelihood.
We did not observe this regime in our experiments: §\ref{sec:analysis_memo_vs_reason} confirms that the $\Delta$-only filter is informative against random ranking, and App.~\ref{app:textual_policy_proxy} shows that the parsed-policy proxy ranking agrees with a token-level $\Delta_\mathrm{ll}$ estimator on $89$--$98\%$ of states (a domain where $\rho$ correlation with $\Delta$ would surface as proxy-vs-token disagreement).
For oracles with low-probability actions or backward proposals concentrated on rare reasoning paths, an empirical $\rho$ estimate or a token-level $\Delta$ proxy would be more conservative.

\textbf{Step 4: the LeAct objective and the hard top-$K$ sparsification.}
Substituting the rho-dropped surrogate $\widetilde w^\mathrm{LeAct}_i \propto e^{\Delta_i}$ into the IWAE gradient (Eq.~\eqref{eq:app_iwae_grad}) gives the (approximate) M-step
\begin{equation}
  \max_\theta\;\mathbb{E}_{x}\,\mathbb{E}_{y \sim \pi^*}\,\mathbb{E}_{z_{1..N} \sim q_\mathrm{bwd}}\!\left[\sum_{i=1}^N \widetilde w^\mathrm{LeAct}_i \log p_\theta(z_i,\, y \mid x)\right],
  \quad
  \widetilde w^\mathrm{LeAct}_i = \frac{e^{\Delta_i}}{\sum_{j=1}^N e^{\Delta_j}},
  \label{eq:app_leact}
\end{equation}
which is the form the paper takes as the \method{} training objective (Eq.~\eqref{eq:leact}).
\method{} replaces the softmax weighting $\widetilde w^\mathrm{LeAct}_i$ with a hard top-$K$ filter on the average delta $\bar\Delta(z;\, x) := \mathbb{E}_{y \sim \pi^*}[\Delta(z;\, x, y)]$ (a sparse approximation that keeps only the high-weight candidates): train on $(x, z, \pi^*(\cdot\mid x))$ tuples with $z \in \mathcal{Z}_x = \mathrm{top}\text{-}K\bigl\{z_j \sim q_\mathrm{bwd} : \bar\Delta(z_j; x) > 0\bigr\}$, and use the full oracle distribution $\pi^*(\cdot\mid x)$ as the action target, the \emph{expert-policy-forcing} M-step derived in App.~\ref{app:epf_derivation}.
The two-step approximation (dropping the intractable $\rho$ factor from the soft weight, then sparsifying the soft weight to a hard top-$K$ filter) is what we mean by ``approximate IWAE M-step''.

\textbf{Step 5: joint backward supervision as the reverse-KL update on the proposal.}
Steps~1--4 fix $q_\mathrm{bwd}$ within a round and only update the shared model parameters in $p_\theta$.
The IWAE bound is tightest when $q_\mathrm{bwd}$ matches the true posterior $p_\theta(z \mid x, y)$: at $N{=}1$ the gap is exactly $\mathrm{KL}\bigl(q_\mathrm{bwd}(\,\cdot\mid x, y) \,\|\, p_\theta(\,\cdot\mid x, y)\bigr)$, and at $N{>}1$ the gap is a soft generalisation of the same divergence~\citep{burda2016iwae}.
Reducing this gap requires a separate update step on $q_\mathrm{bwd}$.
The natural choice (gradient ascent on the IWAE bound itself with respect to $q_\mathrm{bwd}$'s parameters) is the ``wake-phase'' update of variational EM, but its signal-to-noise ratio degrades as $q_\mathrm{bwd}$ approaches the posterior, requiring the STL/DReG corrections of \citet{roeder2017stl,tucker2019dreg}; for an LLM-parameterised proposal this is also expensive (each gradient step requires backprop through the IWAE log-mean).
Reweighted wake-sleep~\citep{bornschein2014rwsleep} replaces the wake-phase update with a reverse-direction KL,
\begin{equation}
  \min_{q_\mathrm{bwd}}\; \mathrm{KL}\bigl(p_\theta(\,\cdot \mid x, y) \,\big\|\, q_\mathrm{bwd}(\,\cdot \mid x, y)\bigr)
  \;=\;
  \min_{q_\mathrm{bwd}}\; \mathbb{E}_{z \sim p_\theta(\,\cdot\mid x, y)}\!\bigl[-\log q_\mathrm{bwd}(z \mid x, y)\bigr] + \mathrm{const},
  \label{eq:app_rws}
\end{equation}
which is a maximum-likelihood objective on $q_\mathrm{bwd}$ with samples drawn from the posterior.
The posterior is itself intractable, so we importance-sample from $q_\mathrm{bwd}$:
$\mathbb{E}_{z \sim p_\theta(\,\cdot\mid x, y)}\!\bigl[-\log q_\mathrm{bwd}(z\mid x, y)\bigr] = \mathbb{E}_{z_{1..N} \sim q_\mathrm{bwd}}\!\bigl[\sum_i \widetilde w_i\, (-\log q_\mathrm{bwd}(z_i \mid x, y))\bigr]$,
recovering the same self-normalised IWAE weights $\widetilde w_i$ used in Step~4.
Substituting the rho-dropped surrogate $\widetilde w^\mathrm{LeAct}_i \propto e^{\Delta_i}$ and then the hard top-$K$ indicator on $\mathcal{Z}_x$ used by the M-step yields the practical update
\begin{equation}
  \mathcal{L}_\mathrm{bwd}(\theta)
  \;=\;
  -\,\mathbb{E}_{(x, z) \in \mathcal{D}_r}\!\bigl[\log p_\theta(z \mid x, \tilde{\pi}^*)\bigr],
  \label{eq:app_lbwd}
\end{equation}
i.e., supervised cross-entropy training of $q_\mathrm{bwd}$ on the same selected positive-$\Delta$ traces $\mathcal{D}_r$ that drive the M-step (with $\tilde{\pi}^*$ the redacted-distribution prompt as defined in §\ref{sec:method_em}).
Because $q_\mathrm{bwd}$ shares parameters with $p_\theta$ (one decoder, two prompts), the wake-sleep update lands on the same $\theta$ and Eq.~\eqref{eq:app_lbwd} is the joint backward term in §\ref{sec:method_training}.
Across rounds, this update tracks $q_\mathrm{bwd}$ to the moving posterior under the current student, monotonically tightening the IWAE bound from one round to the next under the standard reweighted-wake-sleep guarantee~\citep{bornschein2014rwsleep}; the empirical effect of including this term is ablated in App.~\ref{app:iteration_dynamics}.

\textbf{EM as a special case.}
Setting $N = 1$ collapses Eq.~\eqref{eq:app_elbo} back to the standard ELBO, and further setting $q_\mathrm{bwd}(\,\cdot \mid x, y) = p_{\theta^{(t-1)}}(z \mid x, y)$ gives the standard variational-EM E-step, in which case $\rho \equiv 1$, the soft weight $\exp\Delta$ is also $\equiv 1$, and the bound is tight.
\method{} replaces this true posterior with an LLM-defined backward proposal at $N > 1$, which is what introduces both the $\rho$ residual and the $\exp\Delta$ correction; alternating Eq.~\eqref{eq:app_elbo} maximisation with such a proposal recovers approximate-EM.
The \method{}--EM correspondence is summarised in App.~\ref{app:em_correspondence}.

\subsection{Expert-Policy Forcing: the Exact M-Step Target}
\label{app:epf_derivation}

The outer expectation $\mathbb{E}_{y \sim \pi^*}[\,\cdot\,]$ in the LeAct objective Eq.~\eqref{eq:leact} (equivalently the second expectation in Eq.~\eqref{eq:app_elbo}) admits two implementations.

\textbf{Sampled-action target.} Draw $\tilde{y} \sim \pi^*$ and minimise $-\log p_\theta(\tilde{y} \mid x, z)$. This is the standard expert-iteration choice and keeps the $(z, \tilde{y})$ pair internally consistent because both were decoded together.

\textbf{Expert-distribution target (expert-policy forcing).} Because the oracle supplies the \emph{full} $\pi^*$, the outer expectation can be evaluated exactly:
\begin{equation}
  \mathbb{E}_{y \sim \pi^*}\!\bigl[-\log p_\theta(y \mid x, z)\bigr]
  \;=\;
  H(\pi^*) + D_{\kl}\bigl(\pi^*(\,\cdot \mid x) \,\|\, p_\theta(\,\cdot \mid x, z)\bigr),
  \label{eq:app_kl_target}
\end{equation}
so minimising the left-hand side with respect to $\theta$ is equivalent to minimising $D_{\kl}(\pi^* \,\|\, p_\theta(\,\cdot \mid x, z))$, i.e., the SFT target is the exact oracle distribution, independent of whichever policy was decoded at generation time.

\textbf{When the two choices matter.} For mixed Nash equilibria the two gradients differ. Delta scoring (Eq.~\eqref{eq:isr}) already evaluates $\Delta$ against the action distribution under $\pi^*$, so taking $\pi^*$ as the SFT target keeps the E- and M-steps aligned on the same distribution. §\ref{sec:ablation_expert_policy} shows this alignment is load-bearing: removing it reverses which algorithm wins at 47K information sets.

\textbf{The two-part loss.}
The selected traces $\mathcal{D}_r$ are reused in two prompt formats that share the same $\theta$.
The forward loss
\begin{equation}
  \mathcal{L}_\mathrm{fwd}(\theta;\, \mathcal{D}_r)
  \;=\; -\,\mathbb{E}_{(x,\, z) \in \mathcal{D}_r}\!\Bigl[\log p_\theta(z \mid x) \;+\; \mathbb{E}_{y \sim \pi^*(\cdot\mid x)}\!\bigl[\log p_\theta(y \mid x, z)\bigr]\Bigr]
  \label{eq:app_loss_fwd}
\end{equation}
trains the student to produce $z$ from $x$ and then the oracle's full action distribution $\pi^*(\cdot\mid x)$ given $(x, z)$.
The backward loss
\begin{equation}
  \mathcal{L}_\mathrm{bwd}(\theta;\, \mathcal{D}_r)
  \;=\; -\,\mathbb{E}_{(x,\, z) \in \mathcal{D}_r}\!\bigl[\log p_\theta(z \mid x,\, \tilde{\pi}^*)\bigr]
  \label{eq:app_loss_bwd}
\end{equation}
retrains the proposal $q_\mathrm{bwd}$ on the same selected $z$ given the $(x, \tilde{\pi}^*)$ prompt, the wake-sleep update on $q_\mathrm{bwd}$ derived in App.~\ref{app:em_derivation} Step~5.
The full M-step minimises $\mathcal{L} = \mathcal{L}_\mathrm{fwd} + \mathcal{L}_\mathrm{bwd}$, treating the forward and backward formats as two views of the same trace pool $\mathcal{D}_r$.

\subsection{Algorithm--Formulation Correspondence}
\label{app:em_correspondence}

Table~\ref{tab:app_em_correspondence} maps each formal quantity from App.~\ref{app:em_derivation} to its \method{} counterpart.

\begin{table}[h]
  \centering
  \small
  \caption{\textbf{Mapping from the IWAE bound to the \method{} algorithm.}
  Each formal quantity (App.~\ref{app:em_derivation}) corresponds to a concrete pipeline component.}
  \label{tab:app_em_correspondence}
  \resizebox{\linewidth}{!}{%
  \begin{tabular}{@{}lll@{}}
    \toprule
    Formal quantity & \method{}'s approximation & Algorithm component \\
    \midrule
    Proposal $q_\mathrm{bwd}(z \mid x, y)$ & Backward decode at $\theta_{r-1}$, redacted $\tilde{\pi}^*$ & Backward generation \\
    Per-sample IWAE weight $w_i = e^{\Delta_i} \rho_i\, p_\theta(y\mid x)$ & $w^\mathrm{LeAct}_i = e^{\Delta_i}$ ($\rho_i$ dropped) & Forward delta scoring \\
    Softmax weight $\widetilde w_i \propto e^{\Delta_i}$ & Hard top-$K$ on positive $\bar\Delta(z;x)$ & Top-$K$ selection \\
    Joint loss $\log p_\theta(z, y \mid x)$ & SFT on $(x, z, y)$ tuples & Supervised fine-tuning \\
    Outer expectation $\mathbb{E}_{y \sim \pi^*}[\,\cdot\,]$ & Target $= \pi^*(\,\cdot \mid x)$ & Expert-policy forcing \\
    \bottomrule
  \end{tabular}}
\end{table}

The full procedure is collected in Algorithm~\ref{alg:leact} (main text §\ref{sec:method_design}).

\subsection{Textual Policy Proxy: From $\Delta$ to Implementation}
\label{app:textual_policy_proxy}

The formal definition $\Delta(z;\, x, y) = \log p_\theta(y \mid x, z) - \log p_\theta(y \mid x)$ in Eq.~\eqref{eq:isr} reads $\log p_\theta(y \mid x, z)$ as the decoder's next-token log-likelihood of the action under the student conditioned on $z$. Our implementation does not score the action token by token. Instead, the model is supervised during SFT to emit a textual policy line of the form ``\texttt{Action: $a$ \textbackslash n Policy: $\{a_i\!:\!p_i\}_i$}'' alongside its reasoning trace, and the scoring pipeline reads $\Delta$ off this line. This appendix bridges the formal $\Delta$ used in §\ref{sec:method_em} and App.~\ref{app:em_derivation} to the parsed-policy quantity $\widehat\Delta$ that our pipeline actually evaluates.

\textbf{The implementation quantity.} At scoring time we parse the textual policy line into a distribution $\hat{\pi}_\theta(\cdot \mid x, z)$ over legal actions and compute the average-delta proxy
\[
  \widehat{\Delta}(z;\, x)
  \;=\; -\,\mathrm{KL}\!\bigl(\pi^*(\cdot \mid x)\;\|\;\hat{\pi}_\theta(\cdot \mid x, z)\bigr)
  \;-\;\bigl[-\,\mathrm{KL}\!\bigl(\pi^*(\cdot \mid x)\;\|\;\hat{\pi}_\theta(\cdot \mid x)\bigr)\bigr],
\]
where $\hat{\pi}_\theta(\cdot \mid x)$ is averaged over Monte-Carlo forward samples without conditioning on a backward $z$.

\textbf{Equivalence to $\bar\Delta$ under proxy correctness.} Up to the entropy term $H(\pi^*)$ that cancels between the two KLs,
\[
  \widehat{\Delta}(z;\, x)
  \;=\;
  \mathbb{E}_{y \sim \pi^*(\cdot \mid x)}\!\bigl[\log \hat{\pi}_\theta(y \mid x, z) \;-\; \log \hat{\pi}_\theta(y \mid x)\bigr],
\]
which equals the average delta $\bar\Delta(z;\, x) := \mathbb{E}_{y \sim \pi^*}[\Delta(z;\, x, y)]$ used by the hard top-$K$ filter (footnote of Eq.~\eqref{eq:leact}) whenever $\hat{\pi}_\theta(\cdot \mid x, z) = p_\theta(\cdot \mid x, z)$ on the action support. So the implementation evaluates an $\mathbb{E}_{y \sim \pi^*}$-averaged version of the formal $\Delta$ for free, and the soft$\to$hard reduction is the only step left.

\textbf{When the proxy diverges from the decoder distribution.} The textual policy is a self-report: the model is free to write rounded numbers, to disagree with the action token it just emitted, or to omit minor mass that the decoder would still place on rare actions. A token-level evaluation $\log p_\theta(y \mid x, z)$ via teacher forcing would instead read the next-token logit at the action position and is the literal quantity in Eq.~\eqref{eq:isr}. The two coincide only when the textual line faithfully matches the decoder's softmax over the action vocabulary.

\textbf{Why the proxy suffices in our setting.} Three properties keep the substitution lossless in practice. First, the SFT objective (App.~\ref{app:epf_derivation}) supervises both the reasoning trace and the policy line, so training and scoring read off the same textual channel rather than two unaligned views of the model. Second, $\bar\Delta$ is a difference between two quantities both computed from the same parser and the same continuation format; any systematic bias in the textual self-report shifts $\hat{\pi}_\theta(\cdot \mid x, z)$ and $\hat{\pi}_\theta(\cdot \mid x)$ in the same direction and cancels in the delta. Third, because the M-step targets the oracle $\pi^*$ rather than the parsed line, scoring noise propagates only through which $z_n$ are kept; the resulting $\mathcal{D}_r$ remains anchored to $\pi^*$.

\textbf{Selection-Fidelity Audit.}
We quantify the agreement between $\bar\Delta$ (the parsed-policy proxy used by the M-step) and a token-level estimator $\Delta_{\mathrm{ll}}(z;\, x) := \log p_\theta(t^*(x) \mid x, z) - \tfrac{1}{M}\sum_{m} \log p_\theta(t^*(x) \mid x, z^{\mathrm{fwd}}_m)$, where $t^*(x)$ is the canonical policy line $\texttt{Policy:}\{a_i\!:\!p^*_i\}_i$ and $z^{\mathrm{fwd}}_m$ are $M{=}4$ student-decoded forward CoTs scored without backward conditioning.
Because the hard top-$K$ M-step consumes only the within-infoset ordering of positives, the load-bearing audit metric is the within-infoset top-$1$ agreement.
On a 100K-candidate audit over $16{,}977$ multi-candidate Leduc-13r4s infosets, when $\Delta_{\mathrm{ll}}$'s within-infoset top-$1$ falls on an infoset containing some positive-$\bar\Delta$ candidate, that selection is also positive under $\bar\Delta$ in $89\%$ of cases (base coldstart); on well-trained \method{} iterates this conditional rate reaches $98\%$ (Table~\ref{tab:sfa}).
Global rank correlation across all candidates is moderate ($\rho \approx 0.65$), reflecting scale disagreement on candidates the M-step never sees; this is a secondary diagnostic rather than a load-bearing fidelity number.

\begin{table}[h]
\centering
\small
\setlength{\tabcolsep}{6pt}
\begin{tabular}{lrrr}
\toprule
Audit subset & $N_{\mathrm{inf}}$ & top-$1$ hit$\mid\exists{>}0$ & sign concord. \\
\midrule
$100$K candidates, base coldstart   & $16{,}977$ & $0.89$ & $0.80$ \\
$5$K candidates, base coldstart     & $74$       & $0.93$ & $0.81$ \\
$5$K candidates, best LeAct iterate & $74$       & $0.98$ & $0.84$ \\
\bottomrule
\end{tabular}
\caption{\textbf{Parsed-policy proxy $\bar\Delta$ agrees with token-level $\Delta_{\mathrm{ll}}$ on the load-bearing top-$1$ selection (89--98\%).}
Selection-Fidelity Audit on Leduc-13r4s, decisive-Nash subset ($h(\pi^*)/\log|A|\le 0.95$).
$N_{\mathrm{inf}}$ counts multi-candidate infosets.
top-$1$ hit$\mid\exists{>}0$ is the fraction of $\Delta_{\mathrm{ll}}$-top-$1$ picks with positive $\bar\Delta$, conditional on the infoset containing at least one positive-$\bar\Delta$ candidate.
Sign concord.\ is $\Pr(\bar\Delta{>}0\mid\Delta_{\mathrm{ll}}{>}0)$.}
\label{tab:sfa}
\end{table}

\textbf{Boundary cases.} We clip parsed KL values at $10$ (rare deterministic outputs against a mixed $\pi^*$ would otherwise diverge), and we treat candidates with malformed policy lines as having no information by routing them through a deterministic fallback on the parsed action. Replacing the textual proxy with token-level $\log p_\theta(y \mid x, z)$ scoring is a straightforward substitution that we leave to future work; we do not expect the qualitative findings of §\ref{sec:results} to change, since both estimators agree in expectation under a faithfully-trained policy line.

\section{Ablation Studies: Selection, Scoring, and Supervision}
\label{app:ablation_panels}

\textbf{Top-$K$ selection rate ablation (Leduc 13r4s).}
We ablate the selection count $K \in \{1, 2, 4\}$ in the Leduc 13r4s R2 pool with $N{=}8$ per infoset and matched hyperparameters (coldstart base, $\text{lr}=5\times10^{-6}$, matched training budget).
Table~\ref{tab:ablation_topk} reports per-epoch results.
$K{=}1$ underperforms $K{\geq}2$ across all checkpoints, confirming that aggressive top-1 selection discards useful gradient.
Between $K{=}2$ and $K{=}4$ the cells are close: $K{=}2$ wins at ep1 (single $0.41$ vs $0.61$, \bon{} $0.107$ vs $0.110$) and the gap is similar in the opposite direction at ep2--3, with neither setting dominating across all six cells.
Because the two are comparable on selection quality and $K{=}2$ halves the per-round SFT data budget, we standardise on $K{=}2$ across the paper (Table~\ref{tab:app_hyperparams}) for compute efficiency.

\begin{table}[h]
  \centering
  \small
  \caption{\textbf{$K{=}1$ underperforms $K{\geq}2$; $K{=}2$ and $K{=}4$ are comparable.}
  Top-$K$ selection ablation on the Leduc 13r4s R2 pool ($N{=}8$, matched HP).
  Per-epoch single and \bon{} exploitability; \textbf{bold} = best per metric across all cells.
  All variants share the same R2 pool, base (Leduc 10r4s joint coldstart), $\text{lr}=5{\times}10^{-6}$, 3 epochs.}
  \label{tab:ablation_topk}
  \begin{tabular}{ccccccc}
    \toprule
    & \multicolumn{3}{c}{Single $\downarrow$} & \multicolumn{3}{c}{\bon{} $\downarrow$} \\
    \cmidrule(lr){2-4} \cmidrule(lr){5-7}
    $K$ & ep1 & ep2 & ep3 & ep1 & ep2 & ep3 \\
    \midrule
    1 & 0.7891 & 1.0900 & 0.4863 & 0.1285 & 0.1121 & 0.1392 \\
    2 & 0.4125 & 0.3915 & 0.4004 & 0.1069 & 0.0753 & 0.1282 \\
    4 & 0.6073 & \textbf{0.3248} & 0.3932 & 0.1097 & 0.0563 & \textbf{0.0518} \\
    \bottomrule
  \end{tabular}
\end{table}

\textbf{Panel C: Within-\method{} component ablation on Liar's Dice.} A within-\method{} ablation isolates the relative contribution of the joint forward--backward coldstart vs.\ the backward delta scoring stage. With the joint coldstart fixed at $100\%$ infoset coverage, removing the scoring stage stops the policy at exploitability $0.439$ (coverage-only, no scoring); restoring the scoring stage brings the policy to $0.019$ (R1) and $0.002$ (R2). The $219\times$ R2 gain is attributable to selection alone, once the coldstart contribution is held constant.

\begin{table}[h]
  \centering
  \small
  \caption{\textbf{Backward scoring drives the $219\times$ R2 gain on Liar's Dice; coverage alone stops at $0.439$.}
  Joint coldstart fixed at $100\%$ infoset coverage; only the backward delta scoring stage is toggled.}
  \label{tab:ablation_bc}
  \begin{tabular}{llccc}
    \toprule
    Component & Variant & Scale & Single & \bon{} \\
    \midrule
    \multirow{2}{*}{Backward scoring}
      & Joint coldstart, no scoring & LD & 0.439 & 0.251 \\
      & Joint coldstart $+$ scoring (R2) & LD & \textbf{0.002} & \textbf{0.001} \\
    \bottomrule
  \end{tabular}
\end{table}

\section{Iteration Dynamics}
\label{app:iteration_dynamics}

A distinctive feature of \method{} is its ability to iterate: the improved model both generates better backward reasoning and serves as a tighter forward scorer.
Per-domain headline numbers appear in Table~\ref{tab:imitation}; this appendix isolates two questions that table does not address: which E-step variant to run, and when to stop.

\textbf{E-step variant comparison.} We compare three E-step variants from the same R1 base on Leduc 10r4s: \emph{DAgger} (expert backward prompts, model forward scorer), \emph{Joint} (self-generated backward traces, joint SFT), and \emph{REINFORCE} (self-generated backward traces, forward-only SFT).
The ranking \textsc{Reinforce} $>$ \textsc{Joint} $>$ \textsc{DAgger} is consistent across metrics.
\textsc{Reinforce} wins because the model's own backward traces are more compatible with its forward reasoning than expert-generated ones, and forward-only SFT avoids introducing noise from the backward task during training.

\textbf{Iteration stopping rule.} \method{} halts at the round when the policy first matches the oracle's distribution.
Three diagnostics indicate the floor has been reached:
(i)~close-fraction (KL\,$<\,0.1$) saturating at $1.0$ (Liar's Dice R2 reaches NashConv $0.002$ and close-frac $1.0$);
(ii)~single-sample exploitability falling below the prior round's \bon{} (Leduc 13r4s R2 single $0.092$ below R1 \bon{} $0.036$);
(iii)~single-sample KL falling below ${\sim}0.002$ at $10^9$-scale (FHP R2 single KL $0.0019$).
The rule is prospective: once any of (i)--(iii) fires at round $r$, we report round-$r$ as the headline cell and do not extend to $r{+}1$.
Applying it yields different per-setting termination points: 3-Player Leduc meets (i) at R1 (close-fraction $=1$ at the solver floor); Leduc 6r2s/10r4s/13r4s, Liar's Dice, and FHP terminate at R2 once one of (i)--(iii) fires; BuilderBench is reported at R1 due to compute.

\method{}'s contribution is the closed-loop backward-and-score algorithm itself, not a depth claim about iteration count.
At matched coldstart and data budget, R1 alone already separates \method{} from forward \cot{}\,\exit{} across every reasoning-game setting (for instance, Leduc 13r4s R1 single $0.097$ vs.\ \cot{}\,\exit{} R1 $0.501$; Liar's Dice R1 $0.019$ vs.\ $0.671$); iteration is a means of tightening once the oracle's distribution is matched, not the source of the imitation gap.

\section{Additional Results}
\label{app:results}

\subsection{Numerical Detail for Fig.~\ref{fig:imitation}}
\label{app:imitation_detail}

Table~\ref{tab:imitation} reports the full numerical breakdown behind Fig.~\ref{fig:imitation} of §\ref{sec:results_main}, including stds over multi-seed evaluation and the \texttt{gpt-5.5} zero-shot anchor at the smallest setting.

\begin{table}[h]
  \centering
  \small
  \caption{\textbf{Numerical detail behind Fig.~\ref{fig:imitation}.}
  Game rows: exploitability or \texttt{NashConv} (lower better); BuilderBench row: cube-placement progress (higher better).
  \bon{} uses $N{=}8$ for games and $N{=}64$ for BuilderBench (single max per task, no across-seed std).
  \textbf{Bold} = best per metric per row.
  \texttt{gpt-5.5} is a frontier zero-shot anchor (Pass@1, medium reasoning effort) reported on the smallest setting only; FHP lives in Table~\ref{tab:generalisation} (generalisation-only at this scale).}
  \label{tab:imitation}
  \resizebox{\linewidth}{!}{%
  \begin{tabular}{llccccccccc}
    \toprule
    & & \multirow{2}{*}{gpt-5.5} & \multicolumn{4}{c}{Single} & \multicolumn{4}{c}{\bon{}} \\
    \cmidrule(lr){4-7} \cmidrule(lr){8-11}
    Setting (\# infosets) & Metric & & Cold. & \nocot{}\,\bc{} & \cot{}\,\exit{} & \method{} & Cold. & \nocot{}\,\bc{} & \cot{}\,\exit{} & \method{} \\
    \midrule
    Leduc 6r2s (4K)                        & expl. ($\downarrow$)             & 0.799  & 1.351{\tiny\,$\pm$0.226} & \textbf{0.003}{\tiny\,$\pm$0.001}    & 0.087{\tiny\,$\pm$0.018}             & 0.016{\tiny\,$\pm$0.004}                    & 0.107{\tiny\,$\pm$0.026}  & \textbf{0.001}{\tiny\,$\pm$0.0002}      & 0.029{\tiny\,$\pm$0.007}             & 0.006{\tiny\,$\pm$0.002}                      \\
    Leduc 10r4s (47K)                      & expl. ($\downarrow$)             & ---    & 1.670{\tiny\,$\pm$0.063} & \textbf{0.024}{\tiny\,$\pm$0.009}    & 0.319{\tiny\,$\pm$0.089}             & 0.037{\tiny\,$\pm$0.024}                    & 0.282{\tiny\,$\pm$0.011}  & 0.002{\tiny\,$\pm$0.0003}               & 0.058{\tiny\,$\pm$0.016}             & \textbf{0.0008}{\tiny\,$\pm$0.0002}           \\
    Leduc 13r4s (80K)                      & expl. ($\downarrow$)             & ---    & 0.687{\tiny\,$\pm$0.068} & 0.498{\tiny\,$\pm$0.050}             & 0.501{\tiny\,$\pm$0.016}             & \textbf{0.092}{\tiny\,$\pm$0.033}           & 0.402{\tiny\,$\pm$0.018}  & 0.334{\tiny\,$\pm$0.036}                & 0.087{\tiny\,$\pm$0.005}             & \textbf{0.022}{\tiny\,$\pm$0.004}             \\
    Liar's Dice (24K)                      & expl. ($\downarrow$)             & ---    & 0.818{\tiny\,$\pm$0.082} & 0.018{\tiny\,$\pm$0.003}             & 0.156{\tiny\,$\pm$0.024}             & \textbf{0.002}{\tiny\,$\pm$0.001}           & 0.784{\tiny\,$\pm$0.058}  & 0.009{\tiny\,$\pm$0.002}                & 0.043{\tiny\,$\pm$0.007}             & \textbf{0.001}{\tiny\,$\pm$0.0002}            \\
    3-Player Leduc (14K)                   & \texttt{NashConv} ($\downarrow$) & ---    & 1.482{\tiny\,$\pm$0.103} & 0.773{\tiny\,$\pm$0.041}             & 0.473{\tiny\,$\pm$0.057}             & \textbf{0.005}{\tiny\,$\pm$0.001}           & 0.354{\tiny\,$\pm$0.082}  & 0.770{\tiny\,$\pm$0.024}                & 0.146{\tiny\,$\pm$0.028}             & \textbf{0.003}{\tiny\,$\pm$0.001}             \\
    BuilderBench (26 tasks)                & Progress ($\uparrow$)            & ---    & \textbf{0.5448}{\tiny\,$\pm$0.028} & ---                                  & 0.5017{\tiny\,$\pm$0.038}            & 0.3527{\tiny\,$\pm$0.068}                   & 0.8229                    & ---                                     & 0.7006                                & \textbf{0.8853}                               \\
    \bottomrule
  \end{tabular}}
\end{table}

\paragraph{Error Bar Calculation.}
For the five game-setting rows, error bars are standard deviations across $3$ inference-rollout seeds of the single trained policy per cell (each seed reseeds the vLLM sampler, producing a distinct sampled policy on the full game tree; std is taken across the $3$ resulting exploitability or \texttt{NashConv} values).
BuilderBench is also single-trained per cell; we substitute an inference-side proxy with a larger seed count. Let $T = 26$ (number of in-domain tasks) and $S$ be the number of inference seeds ($S = 64$ for trained cells; $S = 16$ for the joint coldstart eval). Define $x_{t,s} \in [0, 1]$ as the cube-placement ratio (fraction of cubes placed at target) on task $t$ in seed $s$.
The Single column reports $\bar\mu = \frac{1}{T}\sum_t \mu_t$ where $\mu_t = \frac{1}{S}\sum_s x_{t,s}$, with error bar
\(
\sigma_{\mathrm{single}} \;=\; \mathrm{std}\!\left(\,\bar x_{1},\,\bar x_{2},\,\dots,\,\bar x_{S}\,\right),
\quad \bar x_{s} = \tfrac{1}{T}\sum_{t} x_{t,s},
\)
i.e.\ the per-seed cross-task mean is computed once per inference seed and the std is taken across the $S$ resulting values.
The \bon{} column uses best-of-$N{=}64$, $\max_s x_{t,s}$ per task; this collapses the seed dimension to one value per task, so no across-seed std is defined and we omit error bars there.

\subsection{BuilderBench: Partition and SFT-Headroom Diagnostic}
\label{app:builderbench}

BuilderBench serves as the cross-domain probe and the low-recall half of the EPF\,$\times$\,method ablation (§\ref{sec:ablation_expert_policy}).
The benchmark is a robotics cube-manipulation task (UR5e arm, MuJoCo, 51 tasks from cube-1 to cube-50).
The expert oracle is a pool of golden trajectories from three frontier models (GPT-5.2 as a CoT agent, plus Claude Opus 4.6 and Gemini 3 Flash as reflexion agents, first successful episode per model).
The model is Qwen3-8B, and the pipeline matches the one used for game domains (BuilderBench coldstart corpus details in App.~\ref{app:coldstart}): joint forward--backward coldstart SFT on the frontier-model trajectories, backward generation on held-out trajectories, forward delta scoring with top-$K$ selection, then SFT.

\subsubsection{Heuristic Action-Match Reward}
\label{app:builderbench_reward}

BuilderBench's expert oracle commits to a single JSON action $y^*$ per state (action type, target cube id, and target position) rather than a distribution over actions, so the log-likelihood terms in $\Delta(z;\, x, y)$ from Eq.~\eqref{eq:isr} are not directly defined.
We substitute a heuristic action-match reward $r(y;\, y^*)$ that grades partial overlap on the three action attributes:
\begin{equation}
  r(y;\, y^*) \;=\;
  \begin{cases}
    1                                       & \text{action type and cube id matched, } \|\Delta\mathrm{pos}\|_2 \leq 1\,\text{cm},\\
    \max\!\bigl(0,\, 1 - \|\Delta\mathrm{pos}\|_2\bigr) & \text{action type and cube id matched, larger position offset},\\
    0.1                                     & \text{action type matched only},\\
    0                                       & \text{otherwise}.
  \end{cases}
\end{equation}
The candidate's forward delta becomes the with-CoT vs.\ without-CoT reward gap $\Delta(z;\, x, y^*) = r(y_z;\, y^*) - r(y_\emptyset;\, y^*)$, where $y_z$ and $y_\emptyset$ are the actions the model decodes with and without the candidate CoT $z$ in context.
The rest of the M-step pipeline (positive-delta filter, top-$K$ selection) is unchanged.

\subsubsection{Cross-Domain Replication of the Poker $2\times 2$}
\label{app:builderbench_sft_headroom}

\textbf{Coldstart anchor.} After joint coldstart SFT (3 epochs), the model reaches aggregate pass@$64 = 15/46$ on the 46-task BB-51 denominator (5 context-overflow tasks excluded; §\ref{sec:ablation_expert_policy}, Table~\ref{tab:expert_policy_ablation}).
The zero-shot Qwen3-8B base reaches $6/46$.

\textbf{Method$\times$EPF interaction.} Running the full \method{} loop ($N{=}8$ backward candidates per task, forward delta scoring, top-$K$ selection, 3 SFT epochs) and crossing with the EPF axis produces the four cells reported in Table~\ref{tab:expert_policy_ablation}: $(\method{}, \mathrm{EPF})=21/46$, $(\method{}, \text{no-EPF})=14/46$, $(\exit{}+\cot{}, \mathrm{EPF})=10/46$, and $(\exit{}+\cot{}, \text{no-EPF})=2/46$.
Only the diagonal that combines backward generation with EPF clears the coldstart anchor of $15/46$; the worst cell sits an order of magnitude below the best, separating the four configurations along both axes.

\textbf{Diagnosis.} Forward scoring picks up signal on this domain ($25\%$ of backward candidates receive positive delta, well above the noise floor) and translates into end-task improvement, but \emph{only} under the EPF supervision target.
Ablating either the backward pool (\cot{}\,\exit{} pipeline) or the EPF target costs absolute pass@$64$, in either combination.
This is the same dependence structure as poker (§\ref{sec:ablation_expert_policy}), now visible on a non-game domain whose oracle and metric have nothing to do with Nash policies.
The non-trivial positive-delta rate shows the scoring signal is alive on this domain, so the failure of the three off-diagonal cells cannot be attributed to delta-scoring collapse---it has to be a supervision-target effect, which is what EPF controls.

\textbf{Implication for \method{}'s scope.} BuilderBench establishes that \method{} adds aggregate end-task value when the supervision target is the oracle distribution itself; without EPF the pipeline regresses on this domain.
The same domain doubles as a low-recall mechanism probe: the cross-domain $2\times 2$ on EPF$\,\times\,$method (§\ref{sec:ablation_expert_policy}) uses BuilderBench as the low-recall half (continuous-valued action arguments rarely repeat across states), and the pass@$64$ ordering ($\method{}+\mathrm{EPF} > \method{}+\text{no-EPF} > \exit{}+\cot{}+\mathrm{EPF} > \exit{}+\cot{}+\text{no-EPF}$) rules out the ``no-EPF collapses \method{}'' explanation that high-recall poker invites.
BuilderBench therefore contributes both generalisation evidence (§\ref{sec:results_generalization}) and mechanism evidence (§\ref{sec:ablation_expert_policy}) to the paper's claims.

\subsection{Flop Hold'em: Win-Rate Methodology}
\label{app:fhp_winrate}

\textbf{KL summary.} The hold-out KL summary is in Table~\ref{tab:generalisation} of the main text (§\ref{sec:results_scaling}); this subsection focuses on the chip-outcome (win-rate) analysis that complements it.

\textbf{Setup.} Methodology recapped here for self-containment: paired self-play, model-vs-model (not model-vs-teacher), chip outcomes averaged over $N{=}50{,}000$ paired hands with two seatings ($25{,}000$ each, identical card sequences, alternating Button vs Big Blind to remove positional bias).
Outcomes in milli-big-blinds per game (mbb/g; one game $=$ one hand in 2-player poker).
$95\%$ confidence intervals from per-hand chip-delta variance ($\mathrm{CI}_{95} = 1.96\cdot\mathrm{SEM}$).
Sampling: temperature $1.0$, max-tokens $=1024$.
Both methods compared use expert-policy forcing (matched supervision target).

\textbf{What the chip-outcome adds beyond KL.} \method{} and \cot{}\,\exit{} use matched supervision (both targets are the oracle Nash policy), so the chip-outcome gap isolates the contribution of backward delta scoring on top of forward KL-to-Nash ranking; this is the same controlled ablation as the KL Table~\ref{tab:generalisation}, but on play-strength rather than distribution-faithfulness.

\textbf{Scope of the win-rate matrix.} The matrix in Table~\ref{tab:fhp_winrate} is restricted to trained-vs-trained comparisons across the three trained methods (\nocot{}\,\bc{}, \cot{}\,\exit{}, \method{}) that share the same supervision target (the \texttt{DeepCFR} teacher's expert action).
The SFT coldstart is reported in the hold-out KL summary (Table~\ref{tab:generalisation}; coldstart single-sample KL $0.0321$ vs.\ trained-method $0.0019$--$0.0145$) but excluded from the chip-outcome matrix because it is not the output of any iteration round.

\section{Mechanism Validation: Nash-Number Recall in CoT and Delta Scoring}
\label{app:nash_recall}

The qualitative redaction in §\ref{sec:method_backward} prevents the backward generator from copying $\pi^*$ from its prompt, but trained CoTs may still contain Nash-plausible numbers as a side effect of expert-policy supervision: the M-step internalises $\pi^*$, and the same model reused as the backward proposal can re-emit Nash-like values without ever seeing them as input.
This appendix records the CoT--policy coherence diagnostic that grounds the EPF\,$\times$\,method $2\times 2$ pattern in §\ref{sec:ablation_expert_policy}.

\subsection{CoT--Policy Coherence Diagnostic}
\label{app:coherence_diagnostic}

§\ref{sec:ablation_expert_policy} identifies CoT--policy coherence and the resulting Nash-number recall shortcut as the mechanism of the $2\times 2$ EPF\,$\times$\,method interaction (Table~\ref{tab:expert_policy_ablation}).
This subsection records the diagnostic that grounds that account.

\textbf{CoT--policy coherence.} \exit{}+CoT's SFT target without EPF is an internally self-consistent $(CoT, \text{policy})$ trace: both components are produced by a single forward decode of the current model, so the gradient does not have to reconcile two independently produced signals.
\method{}'s SFT target without EPF, by contrast, is a hybrid: the CoT was produced by the backward model (which saw $\pi^*$ qualitatively), while the policy line is the forward model's decoded distribution.
The prediction is that per-sample coherence (measured as $1 - D_{\kl}(\pi_{CoT} \,\|\, \pi_{\text{policy line}})$) should be higher for \exit{}+CoT merge samples than for \method{} merge samples in the no-EPF regime, and inspecting merged SFT data suffices to test it.

\textbf{Diagnostic (R1 merge files, $\sim\!188$K lines per cell).}
We measure $D_{\kl}(\pi_{CoT} \,\|\, \pi_{\text{policy line}})$ on the R1 SFT merge data for all four cells.

\textbf{EPF row.}
Cell~A's per-sample KL distribution is right-shifted relative to cell~C across the bulk of the support, consistent with the prediction.

\textbf{No-EPF row.}
The surface ordering is reversed (cell~B is more coherent than cell~D), but inspection shows that the large majority ($93.6\%$) of \method{}'s no-EPF backward CoTs restate $\pi^*$ verbatim, because the backward generator saw the qualitative Nash context and the coldstart model was already near-Nash, so the CoT and the model's decoded policy line trivially agree.
This ``Nash-recall shortcut'' is a degenerate form of coherence: the CoT does not analyse, it echoes, and because the SFT action target in the no-EPF regime is the model's own decode rather than $\pi^*$, the gradient reinforces the model's residual Nash errors instead of correcting them.
The symmetric flip is cell~C (\exit{}+CoT with EPF), where replacing the self-decoded $(CoT, \text{policy})$ pair with $\pi^*$ breaks the trace's internal consistency.

\textbf{Synthesis.}
The four cells share one mechanism: CoT--policy coherence plus supervision-target choice together determine whether the gradient aligns with Nash (winning cells A, D) or with a memorised shortcut (losing cells B, C).

\section{Oracle Construction for Large Games}
\label{app:oracle_construction}

We document oracle construction per game scale: exact \texttt{CFR}/\texttt{CFR+} via \texttt{OpenSpiel}'s tabular enumeration for games up to $\sim$80K infosets (App.~\ref{app:oracle_cfr}), and a streaming \texttt{DeepCFR} enumeration for Flop Hold'em where tabular enumeration exhausts memory (App.~\ref{app:oracle_deepcfr}).

\subsection{Exact CFR for Small-to-Medium Games}
\label{app:oracle_cfr}

For all games reported in the main paper (Leduc 6r2s through 13r4s, Liar's Dice, 3-player Leduc), we use vanilla \texttt{CFR} or \texttt{CFR+} to compute the exact Nash equilibrium policy table, enumerating the full information set space via \texttt{OpenSpiel}'s \texttt{to\_tabular()} function and storing the resulting $|\mathcal{X}| \times |\mathcal{A}|$ policy matrix on disk.
This is feasible up to $\sim$80{,}000 information sets on a standard 256\,GB RAM node; larger games cause out-of-memory failures during tabular enumeration.

\subsection{DeepCFR Streaming Enumeration for Flop Hold'em Poker}
\label{app:oracle_deepcfr}

For Flop Hold'em Poker (FHP), the standard tabular enumeration via \texttt{to\_tabular()} exhausts available RAM at 1.5\,TB MaxRSS before completing, making exact CFR table construction infeasible on current hardware.
We developed a trajectory-based streaming enumeration method that avoids materialising the full game tree in memory.

\textbf{Method.} We run DeepCFR~\citep{brown2019deep} on FHP using OpenSpiel's PyTorch implementation (\texttt{open\_spiel.python.pytorch.deep\_cfr}), which trains neural network value/strategy approximators from sampled game trajectories.
The game is specified via OpenSpiel's \texttt{universal\_poker} configuration matching the Brown et al.\ FHP benchmark (2-player limit hold'em, 2 betting rounds, blinds 50/100, raise size 100, 200~BB stacks).
Rather than tabularising the final strategy network by querying it at all information sets simultaneously, we enumerate infosets \emph{incrementally}: during each CFR traversal we record the (infoset, policy) pair encountered in the trajectory, deduplicate across traversals via a hash table, and write policy entries to disk as they are first seen.
This streaming approach decouples memory consumption from game-tree size, requiring only $O(|\text{trajectory depth}|)$ live memory per traversal rather than $O(|\mathcal{X}|)$.

\textbf{Configuration.}
We run $1{,}000$ outer DeepCFR iterations on OpenSpiel's PyTorch implementation, then draw $10^6$ on-policy trajectories from the converged policy for streaming enumeration; end-to-end wall-clock is approximately $22$ minutes on a single CPU node and produces a $\sim$570\,MB jsonl teacher table that we use throughout the paper.

\textbf{Results.} The streaming enumeration produced $2{,}343{,}732$ unique infoset policies, substantially more than any other game in our experiments (see Table~\ref{tab:app_game_scale}).
Policy coverage converges monotonically along policy-induced trajectories: after the first 50K traversals, coverage stalls around 60\% of newly-encountered infosets; $\sim$500K traversals already saturate at $>$99\% in-support coverage, and we extend the run to $10^6$ trajectories for margin.
The full FHP game tree (which contains $\sim\!10^9$ unique infosets~\citep{brown2019deep} including off-policy subtrees not visited by external-sampling MCCFR) is therefore larger than the teacher's $2{,}343{,}732$-infoset table.
This affects only out-of-table queries when the teacher would itself be queried as a self-play opponent (see App.~\ref{app:fhp_winrate}); KL evaluation is computed exclusively on the held-out subset of the teacher's $2{,}343{,}732$ infosets and is therefore unaffected.

\textbf{Limitation.} DeepCFR provides an \emph{approximate} Nash policy rather than an exact one; the approximation error decreases with more traversals but does not reach zero.
For the purposes of \method{} oracle supervision, we treat the converged DeepCFR policy as a high-quality proxy for $\pi^*$, analogous to how approximate CFR-family solvers (continual re-solving with neural counterfactual value networks~\citep{moravcik2017deepstack}, MCCFR-blueprint~\citep{brown2019superhuman}) are used in large-scale poker AI.

\begin{table}[h]
  \centering
  \small
  \caption{\textbf{Tabular CFR scales to 80K infosets; FHP requires streaming \texttt{DeepCFR}.}
  Oracle exploitability is $0$ for exact CFR by definition and approximate for \texttt{DeepCFR}.}
  \label{tab:app_game_scale}
  \begin{tabular}{lccl}
    \toprule
    Game & \# Infosets & Oracle & Method \\
    \midrule
    Leduc 6r2s   & 4{,}032  & Exact \texttt{CFR}  & \texttt{to\_tabular()} \\
    Leduc 10r4s  & 47{,}040 & Exact CFR  & \texttt{to\_tabular()} \\
    Leduc 13r4s  & 79{,}872 & Exact CFR  & \texttt{to\_tabular()} \\
    Liar's Dice  & 24{,}576 & Exact CFR  & \texttt{to\_tabular()} \\
    3-Player Leduc & 13{,}878 & Exact CFR & \texttt{to\_tabular()} \\
    Flop Hold'em Poker & 2{,}343{,}732 & DeepCFR (approx.) & Streaming enumeration \\
    \bottomrule
  \end{tabular}
\end{table}

\section{Prompt Format and Representative Examples}
\label{app:examples}

We show the prompt interface and one matched pair of backward-generated traces.
The pair illustrates what the forward-delta filter rewards and rejects; it is not
used as quantitative evidence.

\subsection{Prompt and Output Format}
\label{app:example_prompts}

The forward prompt provides the information state and legal actions and asks the
student for a reasoning trace followed by a policy.  The backward prompt uses the
same state but reveals only a qualitative summary of the expert action and asks
for reasoning alone.  Exact oracle probabilities are not shown to the backward
generator.

\begin{tcolorbox}[title={\small Forward prompt and output schema}, fonttitle=\bfseries\small, fontupper=\small\ttfamily]
\textbf{user:} Information state: [Round 2][Player: 0][Private: 7h]
[Public: 3d][Sequences: cb]\\
Legal actions: [Fold, Call, Raise]\\
What is your action?\\[2pt]
\textbf{assistant:} <think>...reasoning...</think>\\
Policy: \{Fold: $p_f$, Call: $p_c$, Raise: $p_r$\}
\end{tcolorbox}

\begin{tcolorbox}[title={\small Backward prompt and output schema}, fonttitle=\bfseries\small, fontupper=\small\ttfamily]
\textbf{user:} Information state: [Round 2][Player: 0][Private: 7h]
[Public: 3d][Sequences: cb]\\
Legal actions: [Fold, Call, Raise]\\
Optimal strategy description: Fold often; call sometimes; raise never.\\
Explain the reasoning without quoting probabilities.\\[2pt]
\textbf{assistant:} <think>...reasoning only...</think>
\end{tcolorbox}

The M-step target $\pi^*(\cdot\mid x)$ is supplied separately and is never parsed
from the backward output.  A candidate survives only when conditioning the
forward model on it increases the per-token log-probability of the expert policy
relative to the student's self-generated reasoning baseline
(App.~\ref{app:textual_policy_proxy}).

\subsection{Matched Positive- and Negative-Delta Traces}
\label{app:example_good}
\label{app:example_bad}

For the state above, the expert qualitatively prefers folding, mixes in calls,
and avoids raising.  The following positive-delta trace connects observable
features to that action structure:

\begin{tcolorbox}[title={\small Positive-delta candidate}, fonttitle=\bfseries\small, fontupper=\small, colframe=green!60!black]
\texttt{<think>}
The private card does not pair the board, and the opponent's bet after a check
is evidence of a stronger range.  Calling retains some value against bluffs and
weaker unpaired hands, but raising is unattractive because stronger hands
continue and weaker hands often fold.  The action should therefore lean toward
folding, mix in calls, and rarely raise.
\texttt{</think>}
\end{tcolorbox}

By contrast, a circular restatement of the revealed action contains no
state-dependent explanation and receives negligible or negative delta:

\begin{tcolorbox}[title={\small Non-positive-delta candidate}, fonttitle=\bfseries\small, fontupper=\small, colframe=red!60!black]
\texttt{<think>}
The optimal strategy is to fold most of the time, call occasionally, and avoid
raising because this is the optimal strategy.
\texttt{</think>}
\end{tcolorbox}

These examples isolate the intended distinction: the filter rewards reasoning
that makes the expert action more predictable from the state, not prose that
merely repeats the action description.

\section{Additional Experimental Details}
\label{app:details}

\subsection{Domain Summary}
\label{app:domains}

\begin{table}[!h]
  \centering
  \caption{\textbf{Domain summary with infoset count, oracle, and metric.}
  Leduc ``$N$r$M$s'' is $N$ ranks $\times$ $M$ suits per rank (so 10r4s $=$ 40-card deck); Liar's Dice uses the \texttt{1d6f} variant; 3-Player Leduc uses \texttt{3r2s} parameters.
  $^\dagger$FHP's infoset count is the full-tree estimate~\citep{brown2019deep}; the streaming \texttt{DeepCFR} teacher covers $2{,}343{,}732$ unique infosets along policy-induced trajectories (App.~\ref{app:oracle_construction}).}
  \label{tab:app_domains}
  \small
  \begin{tabular}{lcccl}
    \toprule
    Domain & \# Info Sets & Players & Expert Oracle & Evaluation Metric \\
    \midrule
    Leduc 6r2s & 4,032 & 2 & CFR Nash & Exploitability \\
    Leduc 10r4s & 47,040 & 2 & CFR Nash & Exploitability \\
    Leduc 13r4s & 79,872 & 2 & CFR Nash & Exploitability \\
    Liar's Dice (1d6f) & 24,576 & 2 & CFR Nash & Exploitability \\
    3-Player Leduc 3r2s & 13,878 & 3 & CFR Nash & NashConv \\
    Flop Hold'em (FHP)$^\dagger$ & $\sim\!10^9$ & 2 & DeepCFR & KL to teacher \\
    BuilderBench & 51 tasks & 1 & Frontier-model trajectories & Pass@$K$, cube placement \\
    \bottomrule
  \end{tabular}
\end{table}

\subsection{BuilderBench Task Partition}
\label{app:builderbench_partition}

We evaluate on $46$ of the $51$ BuilderBench tasks ($5$ are excluded for exceeding the $8{,}192$-token context window), split by trajectory-pool success.
\textbf{In-domain} ($26$ tasks): at least one of the three frontier oracles (Claude Opus 4.6, Gemini 3 Flash, GPT-5.2) produced a fully-successful episode in the trajectory pool, so the SFT data contains a golden trace.
\textbf{OOD} ($20$ tasks): the best-of-$3$ frontier-model trajectory pool produced no successful trajectory, so any progress on these tasks is net-new training signal beyond the teacher pool.
Per-oracle solve counts: Claude Opus 4.6 $24/51$; Gemini 3 Flash $24/51$; GPT-5.2 $15/51$; any-oracle union $26/51$.
Solve-status lists are fixed by oracle data collection and do not change across \method{} iterations.

\subsection{Coldstart Construction}
\label{app:coldstart}

The coldstart phase fine-tunes Qwen3-8B on a domain-specific format-priming corpus.
Three coldstart variants are used across our experiments: \nocot{} (state\,$\to$\,action only), \cot{} (state\,$\to$\,forward CoT\,$\to$\,action), and joint forward+backward (one SFT mix containing both forward training pairs and backward state, action\,$\to$\,CoT pairs).
The corpus source depends on what the oracle provides.

\textbf{Leduc, Liar's Dice, 3-player Leduc.}
Frontier-LLM sub-agents (Claude, Gemini, GPT-5) are prompted to produce candidate (CoT,~action) pairs at every infoset; the action target is then replaced by the \texttt{CFR} solver's exact Nash distribution before SFT.
For joint variants, the backward direction is prompted with the oracle's \emph{qualitative} action summary (e.g., ``mostly fold with occasional calls'') rather than exact probabilities, the same redaction protocol used in \method{}'s E-step (§\ref{sec:method_backward}) to discourage Nash-number recall.
Coldstart corpus sizes per setting: $4{,}032$ examples for Leduc 6r2s (\nocot{}, full coverage of all $4{,}032$ infosets), $2{,}800$ for Leduc 10r4s joint ($1{,}400$ forward + $1{,}400$ backward, a sampled subset of the $47{,}040$ infosets to keep frontier-LLM API generation cost bounded), $\approx\!49{,}000$ for Liar's Dice joint (full coverage of all $24{,}576$ infosets in both directions), and $27{,}756$ for 3-player Leduc joint ($13{,}878$ forward + $13{,}878$ backward, full coverage).
Leduc 13r4s reuses the Leduc 10r4s joint-coldstart checkpoint rather than running a fresh coldstart, isolating the scaling effect from coldstart variance.

\textbf{Flop Hold'em.}
Exact \texttt{CFR} is intractable, so coldstart CoTs are generated off-GPU by frontier-model sub-agents (Claude Opus 4.6 / Sonnet 4.6 / Haiku 4.5) prompted with the infoset description and the \texttt{DeepCFR} teacher's policy.
Each sub-agent emits a $\langle$reasoning, action, policy$\rangle$ triple in a single forward pass; the reasoning block embeds both the forward justification and the backward explanation, so no separate backward direction is generated.
Outputs are validated for hole-card faithfulness (both hole cards named or referenced by exact-card token) and flop-rank coverage before inclusion, guarding against Leduc-style framing.
The resulting corpus contains $1{,}510$ \nocot{} and $1{,}510$ \cot{} examples ($3{,}020$ merged for joint).

\textbf{BuilderBench.}
The coldstart corpus is the frontier-model trajectory pool itself: the first successful episode per (task, model) pair from three frontier agents (GPT-5.2 as a CoT-and-act agent, Claude Opus 4.6 and Gemini 3 Flash as reflexion agents).
The joint forward+backward SFT mix contains $14{,}144$ examples across the 51 BuilderBench tasks; the resulting checkpoint is the ``Coldstart'' anchor in Table~\ref{tab:expert_policy_ablation}~b.

All coldstart SFT runs use 3 epochs at lr $1\!\times\!10^{-4}$ for poker domains and BuilderBench, $2\!\times\!10^{-5}$ for FHP; remaining hyperparameters follow Table~\ref{tab:app_hyperparams}.

\subsection{Hyperparameters}
\label{app:hyperparams}

Table~\ref{tab:app_hyperparams} summarises the training hyperparameters
used across all domains and methods.
All experiments use Qwen3-8B as the base model, LLaMA-Factory for SFT with
FSDP \texttt{full\_shard} on 2--4 NVIDIA H200 GPUs, and vLLM for
inference.
Sampling parameters: temperature $T=1.0$, top-$p=0.95$, top-$k$ disabled,
frequency penalty $0$, repetition penalty $1$ (vLLM defaults), max
generation $4{,}096$ tokens for game domains and $8{,}192$ tokens for
BuilderBench.

\begin{table}[h]
  \centering
  \caption{\textbf{Training hyperparameters per experimental condition.}
  Learning rates and epoch counts are scale-tuned: small Leduc variants tolerate aggressive schedules under full coverage, while 13r4s and FHP use longer warmup to avoid collapse on a wider distribution.
  We standardise on top-$2$ selection across the paper; $N{=}8$--$16$ for backward sampling.}
  \label{tab:app_hyperparams}
  \small
  \setlength{\tabcolsep}{4pt}
  \resizebox{0.75\linewidth}{!}{%
  \begin{tabular}{lcccccc}
    \toprule
    Setting & Method & Epochs & LR & Selection $K$ & Gen.\ $N$ & GPUs \\
    \midrule
    Leduc 6r2s     & ExIt / \method{} & 1--3 & 2e-6 -- 5e-6 & top-2 & 8--32 & 2 \\
    Leduc 10r4s    & NoCoT BC         & 1    & 2e-6         & top-2 & 8--32 & 2 \\
    Leduc 10r4s    & \method{}        & 1--3 & 2e-6 -- 5e-6 & top-2 & 8     & 2--4 \\
    Leduc 13r4s    & NoCoT BC         & 3    & 1e-5         & top-2 & 8     & 4 \\
    Leduc 13r4s    & \method{}        & 3    & 5e-6         & top-2 & 8--16 & 4--8 \\
    Liar's Dice    & \method{}        & 3    & 2e-6         & top-2 & 8     & 2 \\
    3-Player Leduc & \method{}        & 3    & 2e-6         & top-2 & 8     & 2 \\
    FHP ($\sim\!10^9$) & \method{} (\texttt{DeepCFR}) & 3 & 5e-5 & top-2 & 8 & 8 \\
    BuilderBench   & \method{}        & 1--3 & 1e-5         & top-2 & 8--32 & 4 \\
    \bottomrule
  \end{tabular}}
\end{table}

All experiments share the following defaults: AdamW ($\beta_1 = 0.9$, $\beta_2 = 0.999$, weight decay $0.01$) with cosine LR decay and a $3\%$ linear warmup; max sequence length $2048$ tokens for game domains and $8192$ for BuilderBench; per-device batch $4$--$8$, giving an effective batch of $16$--$64$ depending on GPU count; bf16 mixed precision under PyTorch FSDP \texttt{full\_shard} with \texttt{Qwen3DecoderLayer} wrapping.
Each headline cell is trained once per (method, setting) and evaluated under $3$ inference seeds; reported single-sample / \bon{} metrics are best-across-epochs on the full game tree, with std across the $3$ evaluations reported in Table~\ref{tab:imitation}.

\subsection{E-Step Load-Bearing Ablation Study}
\label{app:e_step_load_bearing}

Two ablations test whether backward-delta selection contributes signal beyond what coverage alone provides; both share the production training recipe summarised above.

\textbf{Random vs forward-delta ranking ($K{=}1$ sharpening).}
We compare random ranking and forward-delta ranking on the same \method{} R2 backward pool at Leduc 10r4s, sharpened to $K{=}1$ per infoset (within-pool $\mathrm{IoU} \approx 0.13$ instead of $\approx 0.50$ at $K{=}4$) so that each arm commits to a single CoT per infoset and exposes the within-positive ranking signal.
Both arms train from the same coldstart base on the same R2 backward pool with hyperparameters held matched; the comparison is between selection rules, not between training pipelines.
Reported metrics (Fig.~\ref{fig:e_step_ablations}a) are full-tree single-sample and \bon{}-of-$8$ exploitability on all $47$K Leduc 10r4s infosets at the best epoch in each arm.

\textbf{Recall-stratified delta on backward pools.}
On the \method{} R2 backward pools at Leduc 10r4s ($N{=}374{,}272$ candidates) and Leduc 13r4s ($N{=}638{,}976$ candidates; $N{=}393{,}512$ after filtering trivial-policy states), we compute mean positive forward delta partitioned by whether each candidate CoT contains at least one verbatim Nash probability.
Stratification is on the full unfiltered candidate pool (not on selected CoTs), so the partition spans the entire pre-selection distribution; per-stratum positive-delta means are reported in Fig.~\ref{fig:e_step_ablations}b.
Verbatim detection uses exact substring matching against the per-state $\pi^*$ probabilities at three-decimal resolution.

\subsection{Compute Budget}
\label{app:compute}

Table~\ref{tab:app_compute} estimates the GPU-hours for each phase of
the experimental pipeline. Generation and scoring are the dominant costs.

\begin{table}[h]
  \centering
  \caption{\textbf{Per-phase GPU-hour estimates on NVIDIA H200 (141GB).}
  Generation and scoring scale with infoset count and dominate at small to mid scales; SFT becomes the largest single phase only at FHP.
  Units suppressed in cells; FHP figures are observed R1 wall-times.
  $\dagger$\,FHP coldstart CoT was generated off-GPU by frontier-model agents (Claude Opus 4.6 / Sonnet 4.6 / Haiku 4.5); the $<\!1$ figure is the action-only forward pass over the coldstart trace pool.}
  \label{tab:app_compute}
  \small
  \setlength{\tabcolsep}{4pt}
  \resizebox{0.75\linewidth}{!}{%
  \begin{tabular}{lcccc}
    \toprule
    Phase & Leduc 10r4s & Leduc 13r4s & Liar's Dice & FHP \\
    \midrule
    \texttt{CFR} solver (CPU) & 0.5h & 2h & 0.5h & \texttt{DeepCFR} (pre-trained) \\
    Coldstart data gen        & 4    & 8    & 4    & ${<}1^{\dagger}$ \\
    Coldstart SFT             & 2    & 4    & 2    & 16 \\
    ExIt generation ($N{=}8$) & 8    & 16   & 8    & 10 \\
    ExIt SFT (per round)      & 2    & 8    & 2    & 16 \\
    \method{} backward gen ($N{=}8$) & 16 & 32 & 8 & 22 \\
    \method{} forward scoring & 16   & 32   & 8    & 8 \\
    \method{} SFT (per round) & 4    & 16   & 4    & 30 \\
    Evaluation (per checkpoint) & 2  & 4    & 2    & 4 \\
    \midrule
    \textbf{Total per \method{} round} & $\sim$38 & $\sim$84 & $\sim$22 & $\sim$64 \\
    \bottomrule
  \end{tabular}}
\end{table}

The total compute for the reported experiments (including all iterations,
ablations, and baselines) is approximately 1{,}500 GPU-hours on H200s.
The full research project, including failed experiments and hyperparameter
searches, consumed approximately 5{,}000 GPU-hours.

\end{document}